\documentclass[11pt]{article}

\usepackage[final]{acl}

\usepackage{times}
\usepackage{latexsym}

\usepackage{fontspec}
\usepackage[hindi,english]{babel}

\newfontfamily\hindifont[
  Script=Devanagari,
  Path=./fonts/Noto_Serif_Devanagari/,
]{NotoSerifDevanagari-VariableFont_wdth,wght.ttf}

\newcommand{\hi}[1]{{\hindifont #1}}

\usepackage{microtype}

\usepackage{inconsolata}

\usepackage{graphicx}

\usepackage{fontawesome5}

\usepackage{multirow}
\usepackage{booktabs}
\usepackage{pgfplots}
\pgfplotsset{compat=1.18}
\usepgfplotslibrary{groupplots}

\usepackage[most]{tcolorbox}
\usepackage{tabularx}
\usepackage[table]{xcolor}
\usepackage{amsmath}
\usepackage{subcaption}
\usepackage{amsfonts}
\usepackage{titlesec}
\usepackage{enumitem}
\usepackage{footmisc}
\usepackage{fontspec}

\newfontfamily\emoji[
    Path=./fonts/Noto_Emoji/,
]{NotoEmoji-VariableFont_wght.ttf}

\usepackage{caption}
\definecolor{inputblue}{RGB}{220,230,242}
\definecolor{headerblue}{RGB}{180,200,225}

\tcbset{
  commonstyle/.style={
    boxrule=0pt,
    arc=2pt,
    outer arc=2pt,
    left=6pt,
    right=6pt,
    top=6pt,
    bottom=6pt,
    valign=top
  }
}

\newtcolorbox{inputbox}{
  commonstyle,
  colback=inputblue,
  colframe=headerblue,
  fonttitle=\bfseries,
  title=Input,
}

\newtcolorbox{customoutputbox}[2][]{
  commonstyle,
  colback=inputblue,
  colframe=headerblue,
  fonttitle=\bfseries,
  title=#2,
  fontupper=\small,
  #1
}

\usepackage{listings}
\tcbuselibrary{listings}

\newtcblisting{customoutputcode}[2][]{%
  commonstyle,
  colback=inputblue,
  colframe=headerblue,
  fonttitle=\bfseries,
  title=#2,
  listing only,
  listing options={language=Python,basicstyle=\ttfamily\small,breaklines=true},
  #1
}

\newtcolorbox{findingbox}{
  colback=gray!5,
  colframe=gray!50,
  boxrule=0.8pt,
  arc=2pt,
  left=1pt,
  right=1pt,
  top=1pt,
  bottom=1pt,
  width=\linewidth,
  before skip=3pt,
  after skip=1.5pt,
  fontupper=\textbf{\textit{{\fontsize{10pt}{11pt}\selectfont Finding: }}}\itshape
}

\newcommand{\icon}[1]{%
  
  \raisebox{0.2ex}{%
    \ensuremath{\vcenter{\hbox{\includegraphics[height=1.05em]{#1}}}}%
  }%
}

\title{LLM Parameters for Math Across Languages: Shared or Separate?}

\author{
  Behzad Shomali\textsuperscript{1,2}$^\dagger$ \quad
  Luisa Victor\textsuperscript{2}$^\dagger$ \quad
  Tim Selbach\textsuperscript{2}$^\dagger$ \quad
  \textbf{Ali Hamza Bashir\textsuperscript{1,3}} \\[3pt]
  \textbf{David Berghaus\textsuperscript{1,3}} \quad
  \textbf{Joachim Koehler\textsuperscript{1,3}} \quad
  \textbf{Mehdi Ali\textsuperscript{1,3}} \quad
  \textbf{Markus Frey\textsuperscript{1,2,3}} \\[6pt]
  \textsuperscript{1}Lamarr Institute \quad
  \textsuperscript{2}University of Bonn \quad
  \textsuperscript{3} Fraunhofer IAIS \quad
  $^\dagger$  Equal contribution 
}

\begin{document}

\maketitle
\renewcommand{\thefootnote}{}
\footnotetext{Correspondence to: \texttt{behzad.shomali@uni-bonn.de}}
\renewcommand{\thefootnote}{\arabic{footnote}}

\begin{abstract}

Large language models (LLMs) exhibit substantial cross-lingual variation in mathematical reasoning performance, but it remains unclear whether these differences reflect language-specific parameters or a shared mechanism that manifests differently by language. We present a cross-lingual mechanistic analysis of mathematical reasoning in LLMs, enabling us to localize and compare model parameters that support mathematical reasoning across languages. 
We find that the extracted math-associated parameters exhibit partial cross-lingual overlap, with the strongest overlap concentrated in intermediate model layers.
We further observe that English consistently produces the largest set of math-relevant parameters, whereas lower-resource languages reveal smaller sets of relevant parameters. 
These results suggest that math-related behavior in multilingual LLMs is neither fully language-invariant nor fully language-specific, but instead exhibits partial cross-lingual parameter overlap with systematic language-dependent differences.

\icon{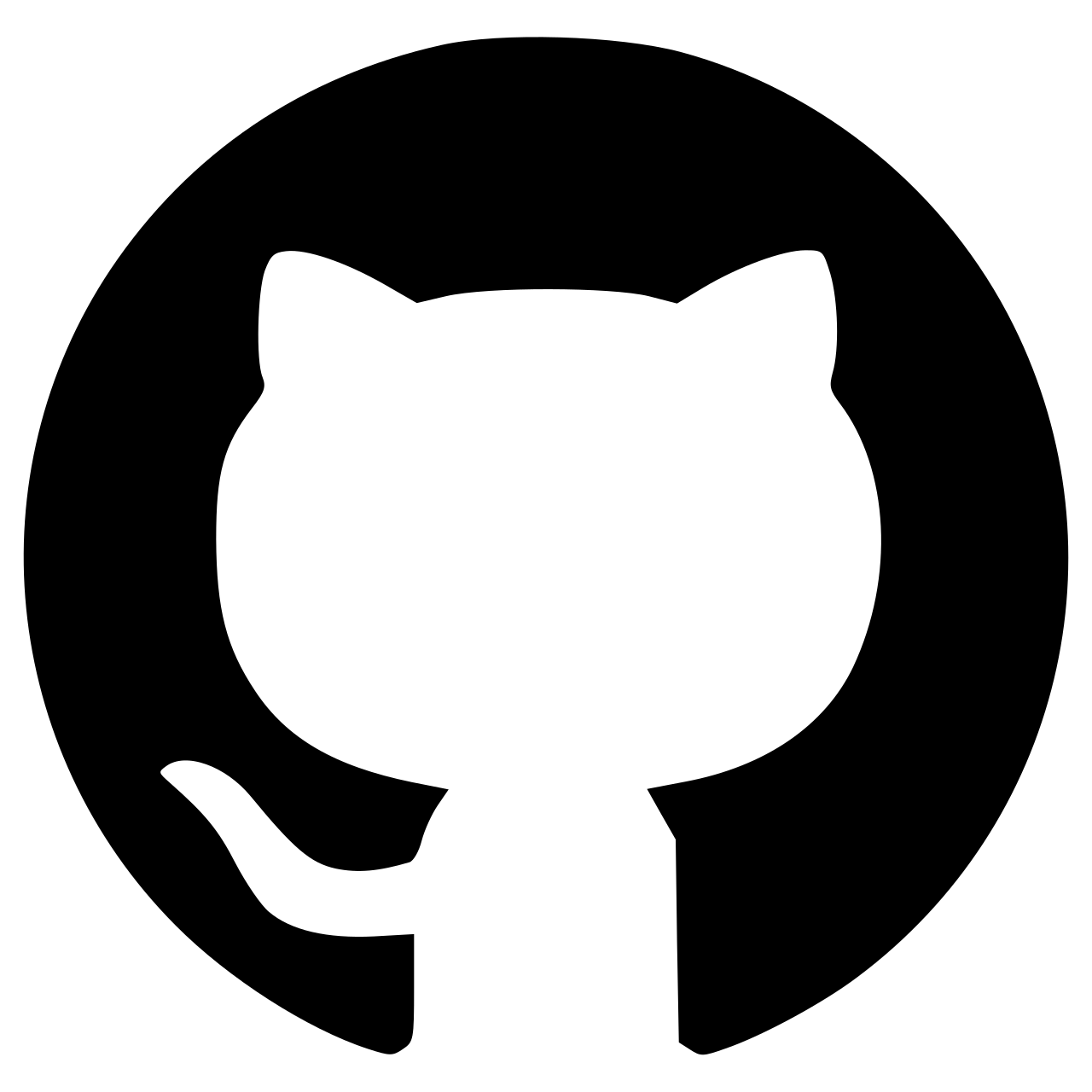}\ \textbf{Code:} \href{https://github.com/luisavictor/math-across-languages}{https://github.com/luisavictor/math-across-languages} 
 
\icon{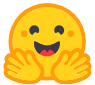}\ \textbf{Translated datasets:}~\href{https://huggingface.co/math-across-languages}{https://huggingface\\.co/math-across-languages}



\end{abstract}


\titlespacing*{\paragraph}{0pt}{0.75ex}{0.75em}

\section{Introduction}

Large language models (LLMs) have achieved remarkable success across a range of tasks, yet their performance on multilingual mathematical reasoning remains uneven \citep{chen2024breaking}. Understanding how these models process mathematical tasks across languages is critical for building robust multilingual systems.

Existing research on LLM capabilities can be broadly categorized into two streams. First, mechanistic studies localize task-specific skills to subsets of model parameters. 
Recent methods leverage forward-pass statistics \citep{sun2023simple}. Notably, \citet{christ2025math} introduced \emph{MathNeurosurgery}, isolating parameters critical to mathematical reasoning in LLMs. Second, multilingual reasoning studies primarily evaluate performance or representational properties. Datasets and strategies like MathOctopus \citep{chen2024breaking} and the UST dataset \citep{ko2025understand} target cross-lingual generalization, while \citet{liu2026large} suggests that English-centric latent pathways may dominate multilingual reasoning. However, these approaches rarely identify whether the same parameters support mathematical reasoning across languages.

Despite these advances, several challenges remain. Mechanistic analyses are mostly limited to English, leaving open questions about cross-lingual parameter sharing. Similarly, performance-focused multilingual studies do not reveal the underlying computational substrates of reasoning. These gaps hinder our understanding of whether mathematical reasoning relies on a universal core or language-specific circuits.
To address this gap, we investigate the localization and overlap of math-specific parameters in LLMs. Our contributions are threefold:

\begin{itemize}[topsep=2pt, partopsep=0pt, itemsep=-0.25ex, leftmargin=1.5em]

    \item \textbf{Mechanistic localization of multilingual math circuits:} We develop a framework to identify and compare model parameters supporting mathematical reasoning across languages.

    \item \textbf{Language-dependent variation in parameter allocation:} We show that English contains the largest set of math-specific parameters, while lower-resource languages have fewer.

    \item \textbf{Functional validation via weight intervention:} We demonstrate that mathematical reasoning emerges as a \textit{collective} property of these localized parameters.

\end{itemize}

\section{Related Work}

\paragraph{Mechanistic analysis of LLMs.} 
Several studies have explored localizing task-specific capabilities within LLM parameters. Gradient-based attribution techniques identify parameters critical to specific tasks \citep{panigrahi2023task}, while forward-pass statistics offer a computationally efficient alternative \citep{sun2023simple}. \citet{christ2025math} proposed \emph{MathNeurosurgery}, which isolates math-specific parameters by comparing weight–activation products on math and non-math datasets. However, their analysis primarily focuses on English and does not examine cross-lingual parameter behavior.

\paragraph{Multilingual LLMs.} 
Research on multilingual LLMs has explored dataset creation and evaluation. MathOctopus \citep{chen2024breaking} and the UST dataset \citep{ko2025understand} target cross-lingual generalization, while other studies suggest that multilingual models often rely on English-centric latent pathways \citep{liu2026large}. These analyses typically evaluate performance or representations, leaving the underlying parameter-level mechanisms unexplored.

\paragraph{Cross-lingual mechanistic studies.} 
Some recent work has investigated neuron activation patterns across translated inputs to identify shared functional units \citep{wang2024sharing}. Our study differs by using a parameter-importance framework, analyzing both attention and Feed-Forward Network (FFN) weights, to localize math-specific circuits and quantify their overlap across languages.
\setlength{\footnotemargin}{11pt}
\titlespacing*{\paragraph}{0pt}{0.75ex}{0.75em}
\setlength{\abovedisplayskip}{4pt}
\setlength{\belowdisplayskip}{4pt}
\setlength{\abovedisplayshortskip}{2pt}
\setlength{\belowdisplayshortskip}{2pt}

\section{Methodology}
We employ the MathNeurosurgery framework~\cite{christ2025math} to identify math-specific parameters and subsequently analyze their overlap, both at the level of the entire model and across individual layers, across multiple languages.



\subsection{Parameter Identification}
Following \cite{christ2025math}, we first compute the task-specific $S_{i j}$ for weight $j$ in neuron $i$: 
\vspace{-2pt}
\begin{equation}
\begin{aligned}
S_{i j} & =\sum_{k=1}^N\left|W_{i j}\right| \cdot\left\|X_j^k\right\|_2 \text { for } X \in \mathcal{D} 
\end{aligned}
\end{equation}

where $W_{ij}$ represents the weight, $|.|$ denotes the absolute value operator, and $\|X_j^k\|_2$ is the L2 norm of the $j$-th feature aggregated across input tokens to normalize the input $X$ from dataset $\mathcal{D}$ for sample $k$. Using this formulation, we compute scores for each parameter over math ($\mathcal{D}_{\text{math}}$) and non-math datasets ($\mathcal{D}_{\text{non-math}}$). 
Then we separately identify the top-$k$ proportion of parameters with the highest score for each task in each layer. Finally, we identify the math-specific parameters by taking the subset of parameters that are crucial for the math task but not for the unrelated task. 



\subsection{Similarity Metric}
To measure similarity between two sets of extracted parameters, we first separately run the previous step for each language, and then measure their pairwise overlaps using the Jaccard coefficient. Let $T_1$ and $T_2$ denote the sets of important parameters representing task-specific parameters for the same task in two languages.
The overlap between the two parameter sets is computed as:
\begin{equation}
\label{eq:jaccard_formula}
J(T_1,T_2) = \frac{|T_1 \cap T_2|}{|T_1 \cup T_2|}
\end{equation}




\titlespacing*{\paragraph}{0pt}{0.6ex}{0.6em}

\section{Results}
We begin with the experimental setup, followed by an analysis of the results.
\subsection{Experimental Setup}
We evaluate Llama~3.2~1B~IT (Llama~1B) \citep{meta2024llama3-2}, Qwen3~4B~IT-2507 (Qwen3~4B) \citep{qwen3technicalreport}, and Llama~3.1~8B~IT (Llama~8B) \citep{meta2024llama3-1} across English, German, French, and Hindi on GSM8K \citep{cobbe2021training} for math reasoning ($\mathcal{D}_{\text{math}}$), MMLU \citep{hendrycks2020measuring} and RACE \citep{lai2017race} as general language understanding datasets ($\mathcal{D}_{\text{non-math}}$). 
We mainly focus on Llama 1B in the main text, while results for larger models are in the Appendices. 


\subsection{Does math performance correlate with the model's capacity for each language?}
\begin{figure}[t]
    \centering
    \includegraphics[width=\linewidth,trim=0 2 0 6,clip]{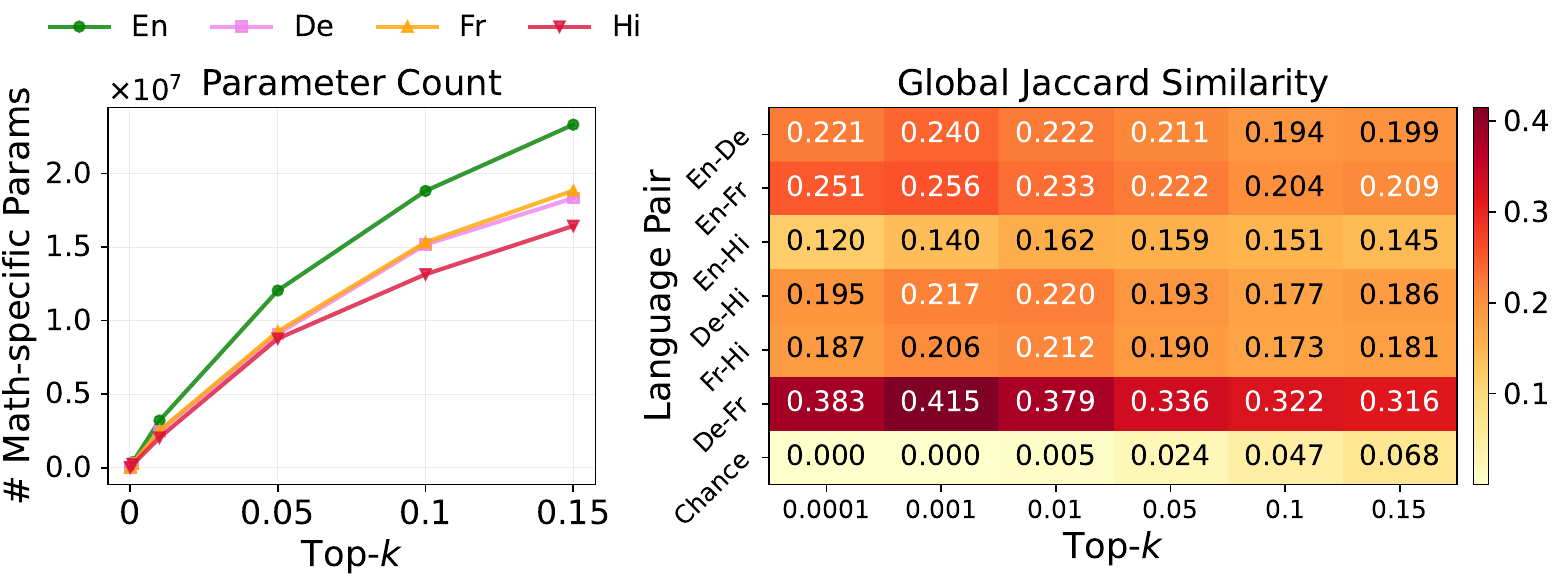}
    \caption{Math-specific parameters in Llama 1B.
    \textbf{Left}: Number of isolated parameters per language.
    \textbf{Right}: Jaccard similarity between language-specific parameter sets.
    }
    \label{fig:overall_jaccard_param_llama1b}
    \vspace{-3pt}
\end{figure}

Figure~\ref{fig:overall_jaccard_param_llama1b} (left) illustrates that, for each top-$k$ value, English consistently exhibits the highest number of math-specific parameters, followed by German and French, and then Hindi. This pattern aligns with the observed reasoning performance across languages, as shown in Table~\ref{tab:pretraining_results}.
More specifically, the trend in the number of math-specific parameters aligns with the relative performance ranking across languages. For example, in Figure~\ref{fig:overall_jaccard_param_llama1b}, at top-$k = 0.01$, English has the highest parameter count, French and German have approximately equal counts, and Hindi has the lowest. Moreover, the gap between Hindi and French is smaller than that between Hindi and English. This pattern becomes even more pronounced with larger models (Figures~\ref{fig:overall_jaccard_param_qwen4b}, \ref{fig:overall_jaccard_param_llama8b}).

\begin{findingbox}
The distribution of math-specific parameters across languages tends to align with their relative reasoning performance in LLMs.
\end{findingbox}

\setlength{\tabcolsep}{4pt}
\begin{table}[t]
    \centering
    \resizebox{\columnwidth}{!}{
    \begin{tabular}{@{}l cc cc cc cc@{}} 
        \toprule
        \multicolumn{1}{l}{\multirow{2}{*}{Model}} & \multicolumn{2}{c}{English} & \multicolumn{2}{c}{German} & \multicolumn{2}{c}{Hindi} & \multicolumn{2}{c}{French} \\ 
        \cmidrule(rl){2-3} \cmidrule(rl){4-5} \cmidrule(rl){6-7} \cmidrule(rl){8-9}
        & GSM8K & RACE & GSM8K & RACE & GSM8K & RACE & GSM8K & RACE \\ \midrule
        Llama 1B & 0.340 & 0.379 & 0.235 & 0.321 & 0.145 & 0.299 & 0.185 & 0.348 \\
        Qwen3 4B & 0.735 & 0.414 & 0.685 & 0.356 & 0.385 & 0.353 & 0.690 & 0.403 \\
        Llama 8B & 0.765 & 0.448 & 0.585 & 0.396 & 0.415 & 0.388 & 0.580 & 0.413\\
 \bottomrule
    \end{tabular}
    }
\caption{Pretraining accuracy (\%) on GSM8K and RACE across English, German, Hindi, and French.
}
\label{tab:pretraining_results}
\vspace{-1pt}
\end{table}




\subsection{Is there a shared cross-lingual core for mathematical reasoning?}
Figure~\ref{fig:overall_jaccard_param_llama1b} highlights the number of math-associated parameters in English (En), German (De), Hindi (Hi) and French (Fr), as well as the Jaccard similarity across language pairs for Llama~1B. Across all language pairs, the Jaccard similarity primarily peaks in the intermediate layers and decreases in deeper layers. The same behavior is also observed in larger models (Figures~\ref{fig:llama1b_all_topk_heatmap}-\ref{fig:Llama-8B_all_topk_heatmap}). This pattern is consistent with prior findings that earlier layers capture more surface-level linguistic features, while later layers can become more specialized and language-dependent \citep{tang2024language}.

Figures~\ref{fig:overall_jaccard_param_llama1b} and~\ref{fig:llama1b_layerwise_dist_math_params_single_topk} demonstrate that English consistently possesses the highest number of math-associated parameters, both in the entire model and in nearly every layer. This implies that the model possesses a more distributed and resilient internal circuit for mathematical computations in English.

Notably, the Jaccard similarity remains substantially above the random baseline across all pairings, except in the initial and final few layers. This suggests the presence of a partially shared parameter substrate associated with mathematical problem solving, especially in intermediate layers. 
Lower similarity in the initial and final layers, especially in larger models (Figures~\ref{fig:Qwen3-4B_all_topk_heatmap},~\ref{fig:Llama-8B_all_topk_heatmap}), likely reflects two distinct phenomena. In early layers, low overlap is consistent with the surface-level processing discussed above. In later layers, it instead points to increased language-specific specialization: with larger parameter budgets, models can mitigate the \emph{curse of multilinguality} \citep{conneau2020unsupervised} by allocating dedicated sub-networks for language-specific reasoning, rather than relying on shared parameters, consistent with other findings on layer specialization in LLMs \citep{frey2026adaptive}.

While Figures~\ref{fig:overall_jaccard_param_llama1b} and~\ref{fig:llama1b_layerwise_jaccard_single_topk} confirm that lexically similar, high-resource languages like French and German exhibit the highest overlap, the lower similarity observed in pairs involving Hindi, particularly En-Hi, suggests a structural barrier imposed by script and tokenization. Because English and Hindi share neither an alphabet nor substantial lexical overlap, the model appears to rely on more disjoint parameter subsets to map divergent input encodings to a shared internal representation. This pattern is also observed in Llama~8B, independent of model scale (Figure~\ref{fig:Llama-8B_all_topk_heatmap}). In contrast, the effect is less pronounced in Qwen3~4B (Figure~\ref{fig:Qwen3-4B_all_topk_heatmap}), potentially reflecting differences in multilingual pretraining coverage, vocabulary design, or tokenization. We emphasize that this interpretation is speculative, as 
our experiments do not isolate these factors.

Surprisingly, the De-Hi and Fr-Hi overlaps in many layers are notably higher than that of En-Hi. We consider two possible explanations. First, French and Hindi as well as German and Hindi have a comparable number of parameters compared to English and Hindi, which explains the lower Jaccard scores for En-Hi. The larger denominator in the similarity computation using Equation~\ref{eq:jaccard_formula} is due to English's larger set of math-associated parameters, which increases the size of the union and reduces the Jaccard similarity. Second, an alternative interpretation is the existence of a shared non-English subspace: a multilingual reasoning manifold that the model exploits when operating outside its primary English-centric pathways. Notably, the pattern in which En-Hi similarity is lower than De-Hi, Fr-Hi, and De-Fr persists across almost all layers and scales to larger models (Figures~\ref{fig:Qwen3-4B_all_topk_heatmap},~\ref{fig:Llama-8B_all_topk_heatmap}). This consistent, surprising observation further motivates the hypothesis that non-English languages may rely on partially overlapping parameter subsets distinct from those used for English.

\begin{figure}[h]
\begin{findingbox}
There exists a partially shared parameter substrate for math problem-solving, which is most prominent in the intermediate layers.
\end{findingbox}
\end{figure}


\input{figures/en_de_hi_llama1B_jaccard}

\subsection{How do weights scaling \& pruning change the model's behavior?}
To validate and investigate the effectiveness of isolated parameters, following \citet{christ2025math}, we apply two interventions: \emph{pruning} and \emph{scaling} (full results in Appendix~\ref{sec:appendix:scaling_pruning_results}). In pruning, weights are set to zero, which ideally impairs mathematical performance more than non-mathematical performance. In scaling, weights are multiplied by a positive factor ($1.1$ for Llama 1B, $1.01$ otherwise). While scaling may improve performance, the effect is less predictable than pruning, as outcomes depend sensitively on the factor choice. In both settings, the intended trend is observed: non-mathematical performance remains largely unchanged, whereas mathematical reasoning is either significantly reduced by pruning or enhanced by scaling.

Analyzing examples where scaling guides the model to the correct answer, we find that the reasoning logic remains largely unchanged, even though the reasoning may not be entirely accurate and may include some missing steps (Figure~\ref{fig:scaling_example_de}). However, for all four languages, the gains mainly come from correcting errors in arithmetic computations, as shown in Figures~\ref{fig:scaling_example_en}-\ref{fig:scaling_example_fr}. This observation explains the trivial improvements observed in larger models, which already have strong arithmetic capabilities and are therefore less responsive to small parameter perturbations.


Tables~\ref{tab:qwen4b_pruning_results_RACE} and \ref{tab:qwen4b_pruning_results_MMLU} demonstrate an emerging pattern when pruning Qwen3~4B model with top-$k$=1e-5. Instead of dropping performance, performance on the English GSM8K task improved remarkably. Manual inspection revealed that suppressing such parameters mainly improved output formatting and better in-context learning (Figure~\ref{fig:qwen_pruning_improvement_example_en}). 
More details in Appendix~\ref{sec:appendix:scaling_pruning_results}.

\begin{findingbox}
Scaling math-specific parameters primarily corrects arithmetic execution rather than reasoning logic.
In some cases, pruning may improve GSM8K accuracy by changing output format or few-shot imitation behavior, rather than by improving reasoning itself.
\end{findingbox}






\subsection{Math-specific parameters effect: collective or individual?}
\label{sec:group_vs_individual}
We investigate whether isolated math-specific parameters exhibit a collective effect, i.e. their impact arises from their aggregation, or an individual effect, where performance changes are driven by a small number of critical parameters.
To this end, for the selected top-$k$ parameters, we sample random subsets at several intermediate proportions. For each proportion, we repeat the sampling process three times and prune the selected parameters. 

\begin{figure}[t]
    \centering
    \includegraphics[width=\linewidth]{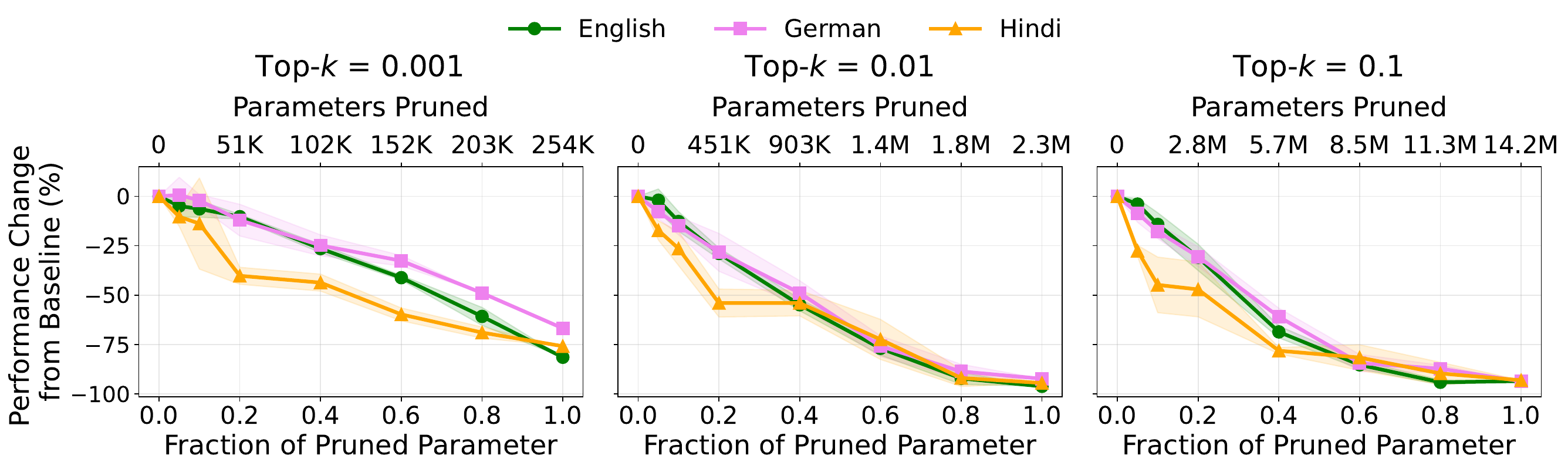}
    \caption{Effect of pruning isolated math-specific parameters on performance across languages for Llama~1B.}
    \label{fig:llama1b_group_vs_individual}
    \vspace{-10pt}
\end{figure}

Figure~\ref{fig:llama1b_group_vs_individual} depicts the results for Llama 1B across three languages.
The observed trend is approximately linear, suggesting that the isolated parameters effect is predominantly \textit{collective} rather than driven by individual parameters. Specifically, the linear relationship indicates that performance degradation scales with the proportion of important parameters removed. Moreover, the low standard deviation supports this conclusion, indicating that the results are not sensitive to the specific subset selected; instead, performance degradation is primarily determined by the proportion of parameters ablated. We observe similar trends for larger models (Figures~\ref{fig:qwen3_group_vs_individual}, \ref{fig:llama8b_group_vs_individual}).

\begin{findingbox}
Identified math-specific parameters have a collective influence on the outcome.
\end{findingbox}

\section{Conclusion}


Our mechanistic analysis suggests that math-associated parameters in multilingual LLMs exhibit partial cross-lingual overlap, with the highest overlap concentrated in intermediate layers. At the same time, deeper layers show more language-specific patterns. We find that English yields the largest set of extracted math-associated parameters, consistent with its stronger performance in our experiments. Lower-resource languages and languages with different scripts, such as Hindi, show less overlap with English, while still sharing some overlap with other non-English languages. These results suggest a distinct pattern of overlap among non-English math-associated parameters. Finally, intervention experiments indicate that the effect of the identified parameters is distributed across the selected set, i.e. exhibiting a collective effect, rather than dominated by a small number of individual parameters.

\section{Acknowledgment}
This research has been funded by the Federal Ministry of Education and Research of Germany
and the state of North-Rhine Westphalia as part of the Lamarr Institute for Machine Learning and
Artificial Intelligence.
\section*{Limitations}
Our work has several limitations. First, despite manual verification, the translated datasets may still lack full fluency and could benefit from stronger translation models. Second, we focus exclusively on mathematical reasoning, leaving open whether similar shared–core and language-dependent patterns extend to other reasoning forms. Third, our analysis considers a limited set of languages and model families, which may restrict the generality of our conclusions across scripts and tokenization schemes. We leave these directions for future work.


    

\bibliography{custom}
\clearpage

\appendix

\setcounter{figure}{0}
\setcounter{table}{0}
\renewcommand{\thefigure}{\thesection.\arabic{figure}}
\renewcommand{\thetable}{\thesection.\arabic{table}}

\section{Experimental Setup}
\label{sec:appendix:experimantal_setup}
In this section, we describe the languages, the dataset curation, and the translation process employed throughout this study.


\paragraph{Languages} To evaluate cross-lingual transferability of mathematical reasoning, we consider English (En), German (De), French (Fr), and Hindi (Hi). This selection enables comparisons across both closely related and more distantly related languages within the Indo-European family. English and German belong to the Germanic branch, while French is a Romance language, and Hindi belongs to the Indo-Aryan branch, following the classification outlined by \citet{beekes2011comparative}.



\paragraph{Dataset curation:} Since MMLU includes a substantial number of math-oriented questions, unlike \citet{christ2025math}, who include \textit{all} available subjects, we apply category-based filtering and remove subject areas that are strongly quantitative, as these can confound our parameter extraction by making parameters associated with mathematical reasoning appear as general language parameters. Concretely, we exclude \textbf{mathematics-related categories} \texttt{(abstract\_algebra, elementary\_mathematics, college\_mathematics, high\_school\_mathematics, high\_school\_statistics)}, 
\textbf{physics categories} \texttt{(conceptual\_physics, college\_physics, high\_school\_physics, astronomy)}, 
\textbf{chemistry categories} \texttt{(college\_chemistry, high\_school\_chemistry)}, \textbf{biology} \texttt{(high\_school\_biology)}, \textbf{computer science} \texttt{(college\_computer\_science, high\_school\_computer\_science)}, and \textbf{quantitatively oriented fields} \texttt{(econometrics, professional\_accounting)}.

\paragraph{Translations:}
We translate all datasets into German, Hindi, and French using the pretrained multilingual machine translation model \texttt{facebook/nllb-200-3.3B}~\citep{costa2022no}. 

For \textsc{GSM8K}, our mathematical reasoning benchmark, we employ a placeholder-based preprocessing step to preserve the integrity of mathematical expressions during translation. Specifically, dataset-specific markers such as $\langle\!\langle \cdot \rangle\!\rangle$ and $\#\#\#\#$ are temporarily replaced with placeholders prior to translation and subsequently restored to their original form.

For \textsc{RACE}, we adopt a sentence-aware chunking strategy to accommodate the substantially longer reading passages compared to \textsc{GSM8K} and \textsc{MMLU}. The articles, questions, and answer options are translated independently, after which we reconstruct the hierarchical reading comprehension structure while strictly preserving the original evaluation schema.

To ensure translation reliability, we manually inspected a random sample of approximately $10\%$ of the translated instances. 





\section{Evaluation}
\label{sec:appendix:evaluation}

\setcounter{figure}{0}
\setcounter{table}{0}
\renewcommand{\thefigure}{\thesection.\arabic{figure}}
\renewcommand{\thetable}{\thesection.\arabic{table}}

To assess model performance before and after weight manipulation, we use the
\textit{EleutherAI Evaluation Harness} framework \citet{eval-harness}.
In addition to English, we evaluate on German, Hindi, and French by implementing custom
\texttt{lm\_eval} tasks for GSM8K, RACE, and MMLU that mirror the original English
task specifications (i.e.~matching dataset splits, preprocessing, and scoring).
For GSM8K, we adopt the \texttt{gsm8k\_cot} configuration, which uses an 8-shot
chain-of-thought prompting format and following prior work \citet{christ2025math, brown2024large}, we use a subset of 200 samples of GSM8K test set for efficacy.
Across languages, we evaluate on the same fixed set of examples per task to ensure comparability of results.
All evaluations are repeated for three different random seeds, and the mean and standard deviations are reported accordingly.
To further facilitate a fair comparison across settings, we increase \texttt{max\_new\_tokens} from its default of 150 to 300 for evaluations on Hindi and French. This adjustment accounts for the possibly higher token counts needed to represent Hindi and French texts compared to math responses in other languages due to script differences. It is worth noting that employing varying values of \texttt{max\_new\_tokens} for each language does not undermine the objectives of our experiments. Our primary focus is on assessing the relative improvements or deteriorations observed in the intervention experiments for each language independently.

\section{Jaccard Random Baseline}
\setcounter{figure}{0}
\setcounter{table}{0}
\renewcommand{\thefigure}{\thesection.\arabic{figure}}
\renewcommand{\thetable}{\thesection.\arabic{table}}
\paragraph{Setup (fixed-size, independent masks)}
Let $U$ be a universe of $N$ parameters.
Each set $S_i \subseteq U$ is sampled \emph{independently and uniformly}
from all subsets of size
\[
|S_i| = m, \qquad m = pN,
\]
where $p \in (0,1)$ and $m$ is assumed to be an integer.
For any fixed parameter $\theta \in U$,
\[
\Pr(\theta \in S_i) = \frac{m}{N} = p,
\qquad
\Pr(\theta \notin S_i) = 1 - p.
\]

\paragraph{Difference sets}
Define the one-sided difference sets
\[
A = S_1 \setminus S_2,
\qquad
B = S_3 \setminus S_4,
\]
where all $S_i$ are independent.

A parameter $\theta$ belongs to $A$ if and only if
$\theta \in S_1$ and $\theta \notin S_2$.
Since $S_1$ and $S_2$ are independent,
\[
\Pr(\theta \in A)
= \Pr(\theta \in S_1)\Pr(\theta \notin S_2)
= p(1 - p).
\]
The same holds for $B$:
\[
\Pr(\theta \in B) = p(1 - p).
\]

\paragraph{Expected Jaccard similarity}
The Jaccard similarity between $A$ and $B$ is
\[
J(A,B) = \frac{|A \cap B|}{|A \cup B|}.
\]

For any fixed $\theta \in U$, membership in $A$ and $B$ depends on
disjoint pairs of sets and is therefore independent. Hence,
\[
\Pr(\theta \in A \cap B) = [p(1-p)]^2,
\]
and
\[
\begin{gathered}
\Pr(\theta \in A \cup B) \\
= \Pr(\theta \in A) + \Pr(\theta \in B) - \Pr(\theta \in A \cap B) \\
= 2p(1-p) - [p(1-p)]^2 .
\end{gathered}
\]

For large $N$, the random set sizes concentrate around their expectations,
and the expected Jaccard similarity is well approximated by the ratio of
expectations:
\[
\mathbb{E}[J(A,B)]
\approx
\frac{[p(1-p)]^2}{2p(1-p) - [p(1-p)]^2}.
\]

Factoring out $p(1-p)$ yields the baseline expression:
\[
\boxed{
\mathbb{E}[J(A,B)]
\approx
\frac{p(1-p)}{2 - p(1-p)}
=
\frac{p(1-p)}{p^2 - p + 2}.
}
\]
It is worth noting that, the approximation arises from replacing
$\mathbb{E}[|A \cap B| / |A \cup B|]$
by
$\mathbb{E}|A \cap B| / \mathbb{E}|A \cup B|$.
This becomes accurate when $N$ is large due to concentration of measure.
\section{Intervention Results}
\label{sec:appendix:scaling_pruning_results}
\setcounter{figure}{0}
\setcounter{table}{0}
\renewcommand{\thefigure}{\thesection.\arabic{figure}}
\renewcommand{\thetable}{\thesection.\arabic{table}}

In the following, we present results for scaling (Tables \ref{tab:llama1b_scaling_results_RACE}-\ref{tab:llama8b_scaling_results_MMLU}) and pruning (Tables \ref{tab:llama1b_pruning_results_RACE}-\ref{tab:llama8b_pruning_results_MMLU}) math-specific parameters across different models and multiple languages. These results serve as proof that the parameter identification for the task and language was performed correctly.
In addition to accuracies on GSM8K, MMLU, and RACE, we also report the \texttt{flexible-extract} GSM8K score from the \texttt{lm\_eval} (\textit{lm-evaluation-harness}) framework. The harness computes GSM8K exact-match accuracy after applying an answer-extraction step: a \textit{strict} extractor that follows the dataset’s canonical output format (e.g., \texttt{The answer is <answer>}) and a more permissive \textit{flexible} extractor that searches the generated text for a number-like substring. We include the flexible metric to reduce sensitivity to output formatting, as our analysis is not focused on instruction-following behavior.

\subsection{Robustness of parameter identification}
Across all settings both in scaling and pruning experiments, the standard deviation across three random seeds remains low. We take this as evidence that the identified parameter subset captures math-reasoning capacity in a stable manner, rather than reflecting idiosyncrasies of a particular sample set. Moreover, all scaling experiments yield highly stable accuracies on \textsc{MMLU} and \textsc{RACE}, indicating that scaling the identified parameters does not measurably affect general language understanding. For pruning, the within-dataset variability is more sensitive to the choice of top-$k$, but remains within a bounded and reasonably stable range. This confirms that we isolate parameter subsets that drive math reasoning while being largely irrelevant to general language understanding, which is of high importance to our parameter comparison and Jaccard similarity interpretations.

\subsection{Pruning results}
Across all settings, increasing top-$k$ leads to a substantial decline in math-reasoning performance, while non-math performance remains comparatively stable on \textsc{MMLU} and \textsc{RACE} (Tables \ref{tab:llama1b_pruning_results_RACE}-\ref{tab:llama8b_pruning_results_MMLU}). This selective degradation is the intended effect and indicates that our parameter identification is successful also across model sizes. 

However, surprisingly, when pruning the Qwen3 model with a top-$k$=1e-5 threshold, corresponding to suppressing approximately 16K parameters, the performance on English GSM8K improves substantially, whereas only a marginal improvement is observed on English GSM8K-Flex shown in Table~\ref{tab:qwen4b_pruning_results_RACE}, which may be a strong signal of wrong output formatting. This counter-intuitive gain from parameter reduction aligns with the framework proposed in \cite{ali2025detecting}, which identifies a class of \textit{prominent but detrimental neurons}. These are highly activated units that, while capturing frequent training patterns, often enforce brittle heuristics that hinder the model's ability to generalize to nuanced reasoning tasks.
Manual inspection reveals that this improvement is only partly due to corrections in arithmetic computation; more importantly, it stems from enhanced in-context instruction following. This manifests as shorter reasoning traces that are less likely to be truncated by the maximum generation length, and it also outputs the required output formats. This stabilization of the output path is consistent with the AdaRAS by \citet{dong2026identifying}, which suggests that \textit{over-active} reasoning circuits can lead to logical redundancy and verbosity. By pruning these circuits, we effectively guide the model toward more direct and stable reasoning paths, thereby better mimicking the reasoning style demonstrated in the provided examples.

Figure~\ref{fig:qwen_pruning_improvement_example_en} illustrates a representative example. Prior to pruning, the model persistently follows its own internal reasoning style, presumably acquired during post-training, rather than adhering to the few-shot exemplars. After suppressing approximately 16K of the most highly activated parameters, the model more consistently aligns its reasoning format with that of the provided demonstrations. This behavior provides empirical support for the theory of \textit{Copy Bias} in In-Context Learning \cite{ali2024mitigating}. According to this work, specific parameters act as \textit{stylistic anchors} that prioritize internal priors over contextual cues; pruning them releases the model from these rigid post-training biases, allowing for greater \textit{behavioral flexibility}.

\subsection{Scaling results}
In contrast to our smallest model (Llama 1B), scaling does not yield significant consistent gains on larger models. Given the strong pruning results, we attribute this mainly to an insufficient scaling factor (1.01) rather than to low-quality parameter identification. Since optimizing this hyperparameter is beyond the scope of this work and does not affect our cross-language comparisons, we leave a systematic search over scaling factors to future work.

However, one can observe a relationship between the model’s general language understanding ability and the effect of scaling on performance, including indications of a saturation effect. For the small Llama 1B model, scaling yields performance gains (in English) of up to 9.7\%. For larger models, performance also improves in some cases, but the changes are much smaller, ranging only from -1.4\% to 2.4\% compared to the baseline. This suggests that scaling is most effective when the model has not yet reached a level of saturated performance. In cases where performance is already high, scaling may be limited because the task-relevant parameters have already been sufficiently optimized, rather than because the task-specific weights were poorly identified. For German, Hindi, and French, we frequently observe that baseline general language understanding performance (e.g., measured by MMLU or RACE) influences the gains obtained from scaling. When language understanding performance is too low, as is often the case for Hindi, scaling tends to produce only small positive or even negative effects. One possible explanation is that insufficient language knowledge prevents the model from identifying the optimal language-specific parameters required for mathematical reasoning. For instance, Qwen3 4B shows consistent improvements from scaling for both German and Hindi; however, the gains are larger for German, which has higher baseline language understanding scores on RACE and MMLU. Overall, these results suggest that a sufficient level of language understanding is beneficial for the scaling approach, while very high baseline performance can lead to saturation effects that limit further improvements through scaling.

\begin{table*}[h]
\centering

\resizebox{\textwidth}{!}{
\begin{tabular}{l ccc ccc ccc ccc}
\toprule
\multicolumn{1}{l}{\multirow{2}{*}{Top-$k$}} & \multicolumn{3}{c}{English} & \multicolumn{3}{c}{German} & \multicolumn{3}{c}{Hindi} & \multicolumn{3}{c}{French} \\
\cmidrule(rl){2-4} \cmidrule(rl){5-7} \cmidrule(rl){8-10} \cmidrule(rl){11-13}
 & GSM8K & GSM8K flex & RACE & GSM8K & GSM8K flex & RACE & GSM8K & GSM8K flex & RACE & GSM8K & GSM8K flex & RACE \\
\hline
0.0 \small(Pre-train) & 0.340 & 0.345 & 0.379 & 0.235 & 0.230 & 0.321 & 0.145 & 0.160 & 0.299 & 0.185 & 0.195 & 0.348\\
\hline
0.000001 & +$0.0_{2.9}$\% & +$0.0_{2.9}$\% & \cellcolor{green!6} +$0.5_{0.0}$\% & \cellcolor{green!7} +$0.9_{0.0}$\% & \cellcolor{green!9} +$1.3_{4.3}$\% & +$0.0_{0.0}$\% & \cellcolor{red!40} -$9.0_{6.9}$\% & \cellcolor{red!34} -$7.5_{6.2}$\% & \cellcolor{red!6} -$0.3_{0.0}$\% & \cellcolor{green!8} +$1.1_{0.0}$\% & +$0.0_{5.1}$\% & +$0.0_{0.0}$\%\\
0.00001 & \cellcolor{green!6} +$0.6_{5.9}$\% & \cellcolor{green!6} +$0.6_{5.8}$\% & \cellcolor{green!7} +$0.8_{0.0}$\% & \cellcolor{red!13} -$2.1_{4.3}$\% & \cellcolor{red!13} -$2.2_{4.3}$\% & \cellcolor{green!6} +$0.3_{0.0}$\% & \cellcolor{red!10} -$1.4_{6.9}$\% & \cellcolor{red!12} -$1.9_{6.2}$\% & \cellcolor{red!6} -$0.3_{0.0}$\% & \cellcolor{green!26} +$6.5_{10.8}$\% & \cellcolor{green!21} +$5.1_{15.4}$\% & \cellcolor{red!6} -$0.3_{0.0}$\%\\
0.0001 & \cellcolor{green!12} +$2.4_{8.8}$\% & \cellcolor{green!12} +$2.3_{8.7}$\% & \cellcolor{green!5} +$0.3_{0.0}$\% & \cellcolor{red!13} -$2.1_{4.3}$\% & \cellcolor{red!8} -$0.9_{4.3}$\% & \cellcolor{red!6} -$0.3_{0.0}$\% & \cellcolor{green!9} +$1.4_{6.9}$\% & +$0.0_{6.2}$\% & \cellcolor{green!6} +$0.3_{0.0}$\% & \cellcolor{green!26} +$6.5_{5.4}$\% & \cellcolor{green!21} +$5.1_{5.1}$\% & +$0.0_{0.0}$\%\\
\bottomrule
\end{tabular}
}
\caption{Relative change in accuracy (\%), averaged over 3 seeds (mean $\pm$ std), for scaling with Llama 1B and RACE.}
\label{tab:llama1b_scaling_results_RACE}
\end{table*}

\begin{table*}[h]
\centering

\resizebox{\textwidth}{!}{
\begin{tabular}{l ccc ccc ccc ccc}
\toprule
\multicolumn{1}{l}{\multirow{2}{*}{Top-$k$}} & \multicolumn{3}{c}{English} & \multicolumn{3}{c}{German} & \multicolumn{3}{c}{Hindi} & \multicolumn{3}{c}{French} \\
\cmidrule(rl){2-4} \cmidrule(rl){5-7} \cmidrule(rl){8-10} \cmidrule(rl){11-13}
 & GSM8K & GSM8K flex & MMLU & GSM8K & GSM8K flex & MMLU & GSM8K & GSM8K flex & MMLU & GSM8K & GSM8K flex & MMLU \\
\hline
0.0 \small(Pre-train) & 0.340 & 0.345 & 0.458 & 0.235 & 0.230 & 0.363 & 0.145 & 0.160 & 0.322 & 0.185 & 0.195 & 0.360\\
\hline
0.000001 & \cellcolor{green!12} +$3.5_{2.9}$\% & \cellcolor{green!12} +$3.5_{2.9}$\% & \cellcolor{green!5} +$0.2_{0.0}$\% & \cellcolor{green!11} +$3.0_{4.3}$\% & \cellcolor{green!12} +$3.5_{4.3}$\% & \cellcolor{green!5} +$0.3_{0.0}$\% & \cellcolor{red!9} -$1.4_{6.9}$\% & \cellcolor{red!14} -$3.1_{6.2}$\% & \cellcolor{green!5} +$0.3_{0.0}$\% & \cellcolor{green!14} +$4.3_{5.4}$\% & \cellcolor{green!8} +$1.5_{5.1}$\% & +$0.0_{0.0}$\%\\
0.00001 & \cellcolor{green!10} +$2.4_{2.9}$\% & \cellcolor{green!10} +$2.3_{2.9}$\% & \cellcolor{green!5} +$0.2_{0.0}$\% & \cellcolor{red!21} -$5.5_{4.3}$\% & \cellcolor{red!17} -$4.3_{4.3}$\% & \cellcolor{green!5} +$0.3_{0.0}$\% & \cellcolor{green!15} +$4.8_{0.0}$\% & \cellcolor{green!11} +$3.1_{6.2}$\% & \cellcolor{green!5} +$0.3_{0.0}$\% & \cellcolor{green!13} +$3.8_{5.4}$\% & \cellcolor{green!10} +$2.6_{0.0}$\% & +$0.0_{0.0}$\%\\
0.0001 & \cellcolor{green!8} +$1.5_{2.9}$\% & \cellcolor{green!8} +$1.4_{2.9}$\% & \cellcolor{green!5} +$0.4_{0.0}$\% & \cellcolor{green!11} +$3.0_{4.3}$\% & \cellcolor{green!11} +$3.0_{4.3}$\% & +$0.0_{0.0}$\% & \cellcolor{red!19} -$4.8_{0.0}$\% & \cellcolor{red!14} -$3.1_{6.2}$\% & \cellcolor{green!5} +$0.3_{0.0}$\% & \cellcolor{green!16} +$5.4_{5.4}$\% & \cellcolor{green!12} +$3.6_{5.1}$\% & +$0.0_{0.0}$\%\\
\bottomrule
\end{tabular}
}
\caption{Relative change in accuracy (\%), averaged over 3 seeds (mean $\pm$ std), for scaling with Llama 1B and MMLU.}
\label{tab:llama1b_scaling_results_MMLU}
\end{table*}

\begin{table*}[h]
\centering

\resizebox{\textwidth}{!}{
\begin{tabular}{l ccc ccc ccc ccc}
\toprule
\multicolumn{1}{l}{\multirow{2}{*}{Top-$k$}} & \multicolumn{3}{c}{English} & \multicolumn{3}{c}{German} & \multicolumn{3}{c}{Hindi} & \multicolumn{3}{c}{French} \\
\cmidrule(rl){2-4} \cmidrule(rl){5-7} \cmidrule(rl){8-10} \cmidrule(rl){11-13}
 & GSM8K & GSM8K flex & RACE & GSM8K & GSM8K flex & RACE & GSM8K & GSM8K flex & RACE & GSM8K & GSM8K flex & RACE \\
\hline
0.0 \small(Pre-train) & 0.735 & 0.815 & 0.414 & 0.685 & 0.400 & 0.356 & 0.385 & 0.460 & 0.353 & 0.690 & 0.615 & 0.403\\
\hline
0.000001 & \cellcolor{green!5} +$0.3_{0.0}$\% & +$0.0_{0.0}$\% & \cellcolor{red!8} -$0.2_{0.0}$\% & \cellcolor{green!13} +$3.9_{1.5}$\% & \cellcolor{green!16} +$5.5_{2.5}$\% & \cellcolor{green!5} +$0.3_{0.0}$\% & \cellcolor{red!11} -$0.5_{0.0}$\% & \cellcolor{red!25} -$1.5_{2.2}$\% & \cellcolor{red!12} -$0.6_{0.0}$\% & \cellcolor{green!9} +$2.2_{1.4}$\% & \cellcolor{green!12} +$3.6_{3.3}$\% & \cellcolor{green!5} +$0.2_{0.0}$\%\\
0.00001 & \cellcolor{red!8} -$0.3_{1.4}$\% & \cellcolor{green!9} +$2.1_{0.0}$\% & +$0.0_{0.0}$\% & \cellcolor{green!13} +$3.9_{1.5}$\% & \cellcolor{green!33} +$13.7_{5.0}$\% & \cellcolor{green!6} +$0.8_{0.0}$\% & \cellcolor{green!10} +$2.6_{0.0}$\% & \cellcolor{green!8} +$1.5_{2.2}$\% & \cellcolor{red!16} -$0.8_{0.0}$\% & \cellcolor{green!8} +$1.7_{1.4}$\% & \cellcolor{green!11} +$2.9_{1.6}$\% & \cellcolor{green!6} +$0.7_{0.0}$\%\\
0.0001 & \cellcolor{green!5} +$0.4_{1.4}$\% & \cellcolor{green!5} +$0.2_{1.2}$\% & \cellcolor{red!14} -$0.7_{0.0}$\% & \cellcolor{green!14} +$4.4_{1.5}$\% & \cellcolor{green!19} +$7.0_{2.5}$\% & \cellcolor{green!7} +$1.1_{0.0}$\% & \cellcolor{green!6} +$0.5_{0.0}$\% & \cellcolor{red!13} -$0.7_{4.3}$\% & \cellcolor{red!20} -$1.1_{0.0}$\% & \cellcolor{green!10} +$2.5_{0.0}$\% & \cellcolor{green!7} +$1.3_{0.0}$\% & \cellcolor{green!6} +$0.5_{0.0}$\%\\
\bottomrule
\end{tabular}
}
\caption{Relative change in accuracy (\%), averaged over 3 seeds (mean $\pm$ std), for scaling with Qwen3 4B and RACE.}

\label{tab:qwen4b_scaling_results_RACE}
\end{table*}

\begin{table*}[h]
\centering
\resizebox{\textwidth}{!}{
\begin{tabular}{l ccc ccc ccc ccc}
\toprule
\multicolumn{1}{l}{\multirow{2}{*}{Top-$k$}} & \multicolumn{3}{c}{English} & \multicolumn{3}{c}{German} & \multicolumn{3}{c}{Hindi} & \multicolumn{3}{c}{French} \\
\cmidrule(rl){2-4} \cmidrule(rl){5-7} \cmidrule(rl){8-10} \cmidrule(rl){11-13}
 & GSM8K & GSM8K flex & MMLU & GSM8K & GSM8K flex & MMLU & GSM8K & GSM8K flex & MMLU & GSM8K & GSM8K flex & MMLU \\
\hline
0.0 \small(Pre-train) & 0.735 & 0.815 & 0.710 & 0.685 & 0.400 & 0.585 & 0.385 & 0.460 & 0.471 & 0.690 & 0.615 & 0.548\\
\hline
0.000001 & \cellcolor{green!9} +$1.8_{1.4}$\% & \cellcolor{green!8} +$1.2_{1.2}$\% & +$0.0_{0.0}$\% & \cellcolor{green!12} +$2.6_{1.5}$\% & \cellcolor{green!29} +$9.2_{7.5}$\% & \cellcolor{red!7} -$0.2_{0.0}$\% & +$0.0_{0.0}$\% & \cellcolor{red!40} -$2.2_{2.2}$\% & \cellcolor{green!5} +$0.2_{0.0}$\% & \cellcolor{green!9} +$1.7_{0.0}$\% & \cellcolor{green!10} +$2.0_{1.6}$\% & \cellcolor{green!5} +$0.2_{0.0}$\%\\
0.00001 & \cellcolor{green!5} +$0.3_{1.4}$\% & +$0.0_{0.0}$\% & +$0.0_{0.0}$\% & \cellcolor{green!14} +$3.4_{1.5}$\% & \cellcolor{green!10} +$2.0_{5.0}$\% & \cellcolor{red!7} -$0.2_{0.0}$\% & \cellcolor{green!8} +$1.3_{2.6}$\% & \cellcolor{green!7} +$1.1_{2.2}$\% & \cellcolor{green!5} +$0.2_{0.0}$\% & \cellcolor{green!12} +$2.9_{1.4}$\% & \cellcolor{green!13} +$3.3_{1.6}$\% & \cellcolor{green!5} +$0.4_{0.0}$\%\\
0.0001 & \cellcolor{green!7} +$1.1_{1.4}$\% & \cellcolor{red!34} -$1.8_{0.0}$\% & +$0.0_{0.0}$\% & \cellcolor{green!16} +$4.1_{1.5}$\% & \cellcolor{green!31} +$10.0_{5.0}$\% & \cellcolor{red!7} -$0.2_{0.0}$\% & \cellcolor{green!6} +$0.5_{0.0}$\% & \cellcolor{red!33} -$1.7_{2.2}$\% & \cellcolor{green!5} +$0.2_{0.0}$\% & \cellcolor{green!10} +$2.2_{1.4}$\% & +$0.0_{1.6}$\% & \cellcolor{green!5} +$0.2_{0.0}$\%\\
\bottomrule
\end{tabular}
}
\caption{Relative change in accuracy (\%), averaged over 3 seeds (mean $\pm$ std), for scaling with Qwen3 4B and MMLU.}

\label{tab:qwen4b_scaling_results_MMLU}
\end{table*}

\begin{table*}[h]
\centering

\resizebox{\textwidth}{!}{
\begin{tabular}{l ccc ccc ccc ccc}
\toprule
\multicolumn{1}{l}{\multirow{2}{*}{Top-$k$}} & \multicolumn{3}{c}{English} & \multicolumn{3}{c}{German} & \multicolumn{3}{c}{Hindi} & \multicolumn{3}{c}{French} \\
\cmidrule(rl){2-4} \cmidrule(rl){5-7} \cmidrule(rl){8-10} \cmidrule(rl){11-13}
 & GSM8K & GSM8K flex & RACE & GSM8K & GSM8K flex & RACE & GSM8K & GSM8K flex & RACE & GSM8K & GSM8K flex & RACE \\
\hline
0.0 \small(Pre-train) & 0.765 & 0.775 & 0.448 & 0.585 & 0.590 & 0.396 & 0.415 & 0.490 & 0.388 & 0.580 & 0.585 & 0.413\\
\hline
0.000001 & \cellcolor{green!11} +$1.0_{0.0}$\% & +$0.0_{1.3}$\% & +$0.0_{0.0}$\% & \cellcolor{green!7} +$0.3_{1.7}$\% & \cellcolor{green!10} +$0.8_{1.7}$\% & \cellcolor{red!11} -$0.5_{0.0}$\% & \cellcolor{green!28} +$3.6_{2.4}$\% & \cellcolor{green!15} +$1.6_{0.0}$\% & +$0.0_{0.0}$\% & \cellcolor{green!18} +$2.1_{1.7}$\% & \cellcolor{green!18} +$2.1_{1.7}$\% & \cellcolor{green!6} +$0.2_{0.0}$\%\\
0.00001 & \cellcolor{green!20} +$2.4_{0.0}$\% & \cellcolor{green!17} +$1.9_{0.0}$\% & \cellcolor{green!6} +$0.2_{0.0}$\% & \cellcolor{green!12} +$1.2_{1.7}$\% & \cellcolor{green!19} +$2.2_{1.7}$\% & \cellcolor{red!8} -$0.3_{0.0}$\% & \cellcolor{green!25} +$3.1_{0.0}$\% & +$0.0_{2.0}$\% & +$0.0_{0.0}$\% & \cellcolor{green!8} +$0.5_{1.7}$\% & \cellcolor{green!7} +$0.3_{3.4}$\% & \cellcolor{green!6} +$0.2_{0.0}$\%\\
0.0001 & +$0.0_{1.3}$\% & \cellcolor{red!8} -$0.3_{1.3}$\% & +$0.0_{0.0}$\% & \cellcolor{green!24} +$2.9_{0.0}$\% & \cellcolor{green!29} +$3.7_{0.0}$\% & \cellcolor{red!8} -$0.3_{0.0}$\% & \cellcolor{green!8} +$0.5_{0.0}$\% & \cellcolor{red!40} -$2.7_{0.0}$\% & \cellcolor{green!6} +$0.3_{0.0}$\% & \cellcolor{green!7} +$0.3_{3.4}$\% & \cellcolor{green!7} +$0.3_{3.4}$\% & \cellcolor{green!8} +$0.5_{0.0}$\%\\
\bottomrule
\end{tabular}
}
\caption{Relative change in accuracy (\%), averaged over 3 seeds (mean $\pm$ std), for scaling with Llama 8B and RACE.}
\label{tab:llama8b_scaling_results_RACE}
\end{table*}

\begin{table*}[h]
\centering
\resizebox{\textwidth}{!}{
\begin{tabular}{l ccc ccc ccc ccc}
\toprule
\multicolumn{1}{l}{\multirow{2}{*}{Top-$k$}} & \multicolumn{3}{c}{English} & \multicolumn{3}{c}{German} & \multicolumn{3}{c}{Hindi} & \multicolumn{3}{c}{French} \\
\cmidrule(rl){2-4} \cmidrule(rl){5-7} \cmidrule(rl){8-10} \cmidrule(rl){11-13}
 & GSM8K & GSM8K flex & MMLU & GSM8K & GSM8K flex & MMLU & GSM8K & GSM8K flex & MMLU & GSM8K & GSM8K flex & MMLU \\
\hline
0.0 \small(Pre-train) & 0.765 & 0.775 & 0.679 & 0.585 & 0.590 & 0.538 & 0.415 & 0.490 & 0.450 & 0.580 & 0.585 & 0.512\\
\hline
0.000001 & \cellcolor{green!20} +$1.6_{0.0}$\% & \cellcolor{green!17} +$1.3_{1.3}$\% & \cellcolor{red!6} -$0.1_{0.0}$\% & \cellcolor{green!16} +$1.2_{1.7}$\% & \cellcolor{green!21} +$1.7_{1.7}$\% & \cellcolor{green!6} +$0.2_{0.0}$\% & \cellcolor{red!10} -$0.5_{2.4}$\% & \cellcolor{red!40} -$3.1_{2.0}$\% & \cellcolor{green!15} +$1.1_{0.0}$\% & \cellcolor{green!10} +$0.5_{0.0}$\% & \cellcolor{green!8} +$0.3_{0.0}$\% & \cellcolor{green!6} +$0.2_{0.0}$\%\\
0.00001 & \cellcolor{green!11} +$0.7_{1.3}$\% & \cellcolor{green!8} +$0.4_{1.3}$\% & \cellcolor{red!6} -$0.1_{0.0}$\% & +$0.0_{1.7}$\% & \cellcolor{green!13} +$0.8_{1.7}$\% & \cellcolor{green!6} +$0.2_{0.0}$\% & \cellcolor{green!9} +$0.5_{2.4}$\% & +$0.0_{2.0}$\% & \cellcolor{green!15} +$1.1_{0.0}$\% & \cellcolor{red!8} -$0.3_{1.7}$\% & \cellcolor{red!8} -$0.3_{1.7}$\% & \cellcolor{green!6} +$0.2_{0.0}$\%\\
0.0001 & \cellcolor{green!15} +$1.0_{0.0}$\% & \cellcolor{green!14} +$1.0_{0.0}$\% & \cellcolor{red!8} -$0.3_{0.0}$\% & \cellcolor{green!26} +$2.2_{0.0}$\% & \cellcolor{green!32} +$2.9_{1.7}$\% & +$0.0_{0.0}$\% & \cellcolor{green!23} +$1.9_{2.4}$\% & \cellcolor{green!14} +$1.0_{2.0}$\% & \cellcolor{green!15} +$1.1_{0.0}$\% & \cellcolor{red!8} -$0.3_{1.7}$\% & \cellcolor{red!8} -$0.3_{1.7}$\% & \cellcolor{green!6} +$0.2_{0.0}$\%\\
\bottomrule
\end{tabular}
}
\caption{Relative change in accuracy (\%), averaged over 3 seeds (mean $\pm$ std), for scaling with Llama 8B and MMLU.}
\label{tab:llama8b_scaling_results_MMLU}

\end{table*}

\begin{table*}[h]
\centering

\resizebox{\textwidth}{!}{
\begin{tabular}{l ccc ccc ccc ccc}
\toprule
\multicolumn{1}{l}{\multirow{2}{*}{Top-$k$}} & \multicolumn{3}{c}{English} & \multicolumn{3}{c}{German} & \multicolumn{3}{c}{Hindi} & \multicolumn{3}{c}{French} \\
\cmidrule(rl){2-4} \cmidrule(rl){5-7} \cmidrule(rl){8-10} \cmidrule(rl){11-13}
 & GSM8K & GSM8K flex & RACE & GSM8K & GSM8K flex & RACE & GSM8K & GSM8K flex & RACE & GSM8K & GSM8K flex & RACE \\
\hline
0.0 \small(Pre-train) & 0.340 & 0.345 & 0.379 & 0.235 & 0.230 & 0.321 & 0.145 & 0.160 & 0.299 & 0.185 & 0.195 & 0.348\\
\hline
0.000001 & \cellcolor{green!40} +$9.7_{2.9}$\% & \cellcolor{green!36} +$8.7_{2.9}$\% & \cellcolor{green!5} +$0.3_{0.0}$\% & \cellcolor{red!8} -$9.4_{4.3}$\% & \cellcolor{red!7} -$6.5_{4.3}$\% & \cellcolor{green!9} +$1.2_{0.0}$\% & \cellcolor{red!8} -$9.0_{6.9}$\% & \cellcolor{red!7} -$8.1_{12.5}$\% & \cellcolor{green!7} +$0.7_{0.0}$\% & \cellcolor{red!6} -$3.8_{5.4}$\% & \cellcolor{red!6} -$4.1_{5.1}$\% & \cellcolor{red!5} -$0.3_{0.0}$\%\\
0.00001 & \cellcolor{red!7} -$7.9_{2.9}$\% & \cellcolor{red!8} -$9.3_{2.9}$\% & \cellcolor{green!10} +$1.6_{2.6}$\% & \cellcolor{red!11} -$18.3_{12.8}$\% & \cellcolor{red!10} -$16.1_{13.0}$\% & +$0.0_{0.0}$\% & \cellcolor{red!16} -$31.0_{20.7}$\% & \cellcolor{red!15} -$29.4_{18.8}$\% & \cellcolor{green!12} +$2.0_{0.0}$\% & \cellcolor{red!7} -$7.0_{5.4}$\% & \cellcolor{red!8} -$8.7_{0.0}$\% & \cellcolor{red!5} -$1.4_{2.9}$\%\\
0.0001 & \cellcolor{red!13} -$25.0_{2.9}$\% & \cellcolor{red!14} -$26.1_{2.9}$\% & \cellcolor{red!5} -$0.5_{0.0}$\% & \cellcolor{red!16} -$31.1_{4.3}$\% & \cellcolor{red!15} -$29.6_{4.3}$\% & \cellcolor{green!15} +$2.8_{3.1}$\% & \cellcolor{red!20} -$43.4_{6.9}$\% & \cellcolor{red!16} -$31.2_{18.8}$\% & \cellcolor{red!5} -$1.0_{0.0}$\% & \cellcolor{red!11} -$17.3_{5.4}$\% & \cellcolor{red!11} -$19.0_{5.1}$\% & \cellcolor{red!5} -$0.9_{0.0}$\%\\
\bottomrule
\end{tabular}
}
\caption{Relative change in accuracy (\%), averaged over 3 seeds (mean $\pm$ std), for pruning with Llama 1B and RACE.}
\label{tab:llama1b_pruning_results_RACE}

\end{table*}

\begin{table*}[h]
\centering

\resizebox{\textwidth}{!}{
\begin{tabular}{l ccc ccc ccc ccc}
\toprule
\multicolumn{1}{l}{\multirow{2}{*}{Top-$k$}} & \multicolumn{3}{c}{English} & \multicolumn{3}{c}{German} & \multicolumn{3}{c}{Hindi} & \multicolumn{3}{c}{French} \\
\cmidrule(rl){2-4} \cmidrule(rl){5-7} \cmidrule(rl){8-10} \cmidrule(rl){11-13}
 & GSM8K & GSM8K flex & MMLU & GSM8K & GSM8K flex & MMLU & GSM8K & GSM8K flex & MMLU & GSM8K & GSM8K flex & MMLU \\
\hline
0.0 \small(Pre-train) & 0.340 & 0.345 & 0.458 & 0.235 & 0.230 & 0.363 & 0.145 & 0.160 & 0.322 & 0.185 & 0.195 & 0.360\\
\hline
0.000001 & \cellcolor{green!40} +$3.8_{8.8}$\% & \cellcolor{green!31} +$2.9_{8.7}$\% & \cellcolor{red!5} -$0.2_{0.0}$\% & \cellcolor{red!8} -$9.4_{4.3}$\% & \cellcolor{red!8} -$8.7_{4.3}$\% & \cellcolor{red!5} -$0.6_{0.0}$\% & \cellcolor{red!10} -$15.9_{6.9}$\% & \cellcolor{red!11} -$18.8_{6.2}$\% & \cellcolor{red!5} -$1.2_{0.0}$\% & +$0.0_{5.4}$\% & \cellcolor{red!6} -$3.6_{5.1}$\% & \cellcolor{red!5} -$1.1_{0.0}$\%\\
0.00001 & \cellcolor{red!10} -$14.7_{2.9}$\% & \cellcolor{red!10} -$15.4_{0.0}$\% & \cellcolor{red!6} -$3.7_{0.0}$\% & \cellcolor{red!11} -$18.3_{12.8}$\% & \cellcolor{red!10} -$15.2_{13.0}$\% & \cellcolor{red!5} -$0.8_{0.0}$\% & \cellcolor{red!17} -$35.9_{6.9}$\% & \cellcolor{red!15} -$30.0_{6.2}$\% & \cellcolor{red!5} -$0.3_{0.0}$\% & \cellcolor{red!11} -$18.9_{5.4}$\% & \cellcolor{red!12} -$21.5_{5.1}$\% & \cellcolor{red!6} -$4.7_{2.8}$\%\\
0.0001 & \cellcolor{red!11} -$19.7_{8.8}$\% & \cellcolor{red!12} -$20.9_{8.7}$\% & \cellcolor{red!5} -$2.4_{0.0}$\% & \cellcolor{red!17} -$34.0_{4.3}$\% & \cellcolor{red!16} -$33.5_{4.3}$\% & \cellcolor{red!5} -$0.6_{0.0}$\% & \cellcolor{red!17} -$35.9_{13.8}$\% & \cellcolor{red!15} -$29.4_{6.2}$\% & \cellcolor{red!5} -$1.9_{0.0}$\% & \cellcolor{red!9} -$11.9_{5.4}$\% & \cellcolor{red!10} -$15.4_{5.1}$\% & \cellcolor{red!5} -$0.6_{0.0}$\%\\
\bottomrule

\end{tabular}
}
\caption{Relative change in accuracy (\%), averaged over 3 seeds (mean $\pm$ std), for pruning with Llama 1B and MMLU.}
\label{tab:llama1b_pruning_results_MMLU}
\end{table*}

\begin{table*}[h]
\centering

\label{tab:accuracy}
\resizebox{\textwidth}{!}{
\begin{tabular}{l ccc ccc ccc ccc}
\toprule
\multicolumn{1}{l}{\multirow{2}{*}{Top-$k$}} & \multicolumn{3}{c}{English} & \multicolumn{3}{c}{German} & \multicolumn{3}{c}{Hindi} & \multicolumn{3}{c}{French} \\
\cmidrule(rl){2-4} \cmidrule(rl){5-7} \cmidrule(rl){8-10} \cmidrule(rl){11-13}
 & GSM8K & GSM8K flex & RACE & GSM8K & GSM8K flex & RACE & GSM8K & GSM8K flex & RACE & GSM8K & GSM8K flex & RACE \\
\hline
0.0 \small(Pre-train) & 0.735 & 0.815 & 0.414 & 0.685 & 0.400 & 0.356 & 0.385 & 0.460 & 0.353 & 0.690 & 0.615 & 0.403\\
\hline
0.000001 & \cellcolor{red!12} -$21.8_{9.5}$\% & \cellcolor{red!7} -$6.1_{3.7}$\% & \cellcolor{green!6} +$1.0_{0.0}$\% & \cellcolor{red!5} -$0.3_{2.9}$\% & \cellcolor{green!11} +$4.5_{2.5}$\% & \cellcolor{red!5} -$0.3_{0.0}$\% & \cellcolor{green!17} +$8.6_{2.6}$\% & \cellcolor{red!7} -$7.0_{19.6}$\% & \cellcolor{red!6} -$3.7_{0.0}$\% & \cellcolor{green!10} +$4.1_{2.9}$\% & \cellcolor{green!21} +$11.4_{3.3}$\% & \cellcolor{red!5} -$0.5_{2.5}$\%\\
0.00001 & \cellcolor{green!32} +$19.3_{4.1}$\% & \cellcolor{green!11} +$4.7_{2.5}$\% & \cellcolor{red!5} -$0.7_{0.0}$\% & \cellcolor{red!6} -$3.2_{1.5}$\% & \cellcolor{green!40} +$24.2_{27.5}$\% & \cellcolor{red!5} -$1.1_{0.0}$\% & \cellcolor{red!5} -$1.8_{7.8}$\% & \cellcolor{red!13} -$24.3_{10.9}$\% & \cellcolor{red!6} -$3.7_{0.0}$\% & \cellcolor{green!9} +$2.9_{1.4}$\% & \cellcolor{green!10} +$3.6_{9.8}$\% & \cellcolor{red!6} -$5.2_{0.0}$\%\\
0.0001 & \cellcolor{green!6} +$1.4_{6.8}$\% & \cellcolor{red!10} -$15.7_{8.6}$\% & \cellcolor{red!5} -$2.7_{0.0}$\% & \cellcolor{red!13} -$22.6_{4.4}$\% & \cellcolor{red!14} -$28.0_{27.5}$\% & \cellcolor{red!5} -$0.3_{0.0}$\% & \cellcolor{red!11} -$19.0_{7.8}$\% & \cellcolor{red!16} -$31.5_{23.9}$\% & \cellcolor{red!7} -$7.4_{2.8}$\% & \cellcolor{red!10} -$16.4_{5.8}$\% & \cellcolor{red!16} -$32.8_{26.0}$\% & \cellcolor{red!6} -$3.7_{2.5}$\%\\
\bottomrule
\end{tabular}
}
\caption{Relative change in accuracy (\%), averaged over 3 seeds (mean $\pm$ std), for pruning with Qwen3 4B and RACE.}
\label{tab:qwen4b_pruning_results_RACE}
\end{table*}

\begin{table*}[h]
\centering

\resizebox{\textwidth}{!}{
\begin{tabular}{l ccc ccc ccc ccc}
\toprule
\multicolumn{1}{l}{\multirow{2}{*}{Top-$k$}} & \multicolumn{3}{c}{English} & \multicolumn{3}{c}{German} & \multicolumn{3}{c}{Hindi} & \multicolumn{3}{c}{French} \\
\cmidrule(rl){2-4} \cmidrule(rl){5-7} \cmidrule(rl){8-10} \cmidrule(rl){11-13}
 & GSM8K & GSM8K flex & MMLU & GSM8K & GSM8K flex & MMLU & GSM8K & GSM8K flex & MMLU & GSM8K & GSM8K flex & MMLU \\
\hline
0.0 \small(Pre-train) & 0.735 & 0.815 & 0.710 & 0.685 & 0.400 & 0.585 & 0.385 & 0.460 & 0.471 & 0.690 & 0.615 & 0.548\\
\hline
0.000001 & \cellcolor{red!7} -$6.8_{5.4}$\% & \cellcolor{green!5} +$0.2_{2.5}$\% & \cellcolor{red!5} -$0.7_{0.0}$\% & \cellcolor{green!8} +$3.9_{2.9}$\% & \cellcolor{green!40} +$38.8_{12.5}$\% & \cellcolor{red!5} -$0.7_{0.0}$\% & \cellcolor{green!11} +$7.3_{2.6}$\% & \cellcolor{red!6} -$5.4_{4.3}$\% & \cellcolor{red!6} -$3.4_{0.0}$\% & \cellcolor{green!7} +$2.6_{1.4}$\% & \cellcolor{green!11} +$7.0_{3.3}$\% & +$0.0_{0.0}$\%\\
0.00001 & \cellcolor{green!12} +$8.8_{4.1}$\% & \cellcolor{red!9} -$13.3_{8.6}$\% & \cellcolor{red!5} -$0.8_{0.0}$\% & \cellcolor{green!5} +$0.7_{1.5}$\% & \cellcolor{red!5} -$0.5_{15.0}$\% & \cellcolor{red!5} -$2.4_{1.7}$\% & \cellcolor{green!11} +$7.0_{5.2}$\% & \cellcolor{red!24} -$53.9_{2.2}$\% & \cellcolor{red!5} -$2.5_{0.0}$\% & \cellcolor{green!10} +$6.1_{1.4}$\% & \cellcolor{green!14} +$10.6_{3.3}$\% & \cellcolor{red!5} -$0.2_{0.0}$\%\\
0.0001 & \cellcolor{red!10} -$14.7_{6.8}$\% & \cellcolor{red!18} -$39.0_{3.7}$\% & \cellcolor{red!5} -$1.0_{0.0}$\% & \cellcolor{red!9} -$11.4_{0.0}$\% & \cellcolor{red!11} -$18.3_{15.0}$\% & \cellcolor{red!5} -$2.2_{0.0}$\% & \cellcolor{red!10} -$15.1_{2.6}$\% & \cellcolor{green!9} +$5.0_{4.3}$\% & \cellcolor{red!8} -$9.3_{2.1}$\% & \cellcolor{red!7} -$7.5_{1.4}$\% & \cellcolor{red!11} -$17.4_{16.3}$\% & \cellcolor{red!5} -$0.4_{0.0}$\%\\
\bottomrule
\end{tabular}
}
\caption{Relative change in accuracy (\%), averaged over 3 seeds (mean $\pm$ std), for pruning with Qwen3 4B and MMLU.}
\label{tab:qwen4b_pruning_results_MMLU}
\end{table*}

\begin{table*}[h]
\centering

\resizebox{\textwidth}{!}{
\begin{tabular}{l ccc ccc ccc ccc}
\toprule
\multicolumn{1}{l}{\multirow{2}{*}{Top-$k$}} & \multicolumn{3}{c}{English} & \multicolumn{3}{c}{German} & \multicolumn{3}{c}{Hindi} & \multicolumn{3}{c}{French} \\
\cmidrule(rl){2-4} \cmidrule(rl){5-7} \cmidrule(rl){8-10} \cmidrule(rl){11-13}
 & GSM8K & GSM8K flex & RACE & GSM8K & GSM8K flex & RACE & GSM8K & GSM8K flex & RACE & GSM8K & GSM8K flex & RACE \\
\hline
0.0 \small(Pre-train) & 0.765 & 0.775 & 0.448 & 0.585 & 0.590 & 0.396 & 0.415 & 0.490 & 0.388 & 0.580 & 0.585 & 0.413\\
\hline
0.000001 & \cellcolor{red!5} -$2.4_{3.9}$\% & \cellcolor{red!6} -$3.2_{3.9}$\% & \cellcolor{green!8} +$0.2_{0.0}$\% & \cellcolor{green!40} +$2.1_{1.7}$\% & \cellcolor{green!33} +$1.7_{1.7}$\% & \cellcolor{green!9} +$0.3_{0.0}$\% & \cellcolor{red!8} -$8.9_{14.5}$\% & \cellcolor{red!5} -$2.4_{6.1}$\% & \cellcolor{green!22} +$1.0_{0.0}$\% & \cellcolor{red!7} -$7.4_{6.9}$\% & \cellcolor{red!8} -$10.3_{8.5}$\% & \cellcolor{green!9} +$0.2_{0.0}$\%\\
0.00001 & \cellcolor{green!9} +$0.3_{1.3}$\% & \cellcolor{red!5} -$0.6_{2.6}$\% & \cellcolor{green!35} +$1.8_{0.0}$\% & \cellcolor{red!6} -$4.8_{1.7}$\% & \cellcolor{red!7} -$5.9_{1.7}$\% & \cellcolor{green!17} +$0.8_{0.0}$\% & \cellcolor{red!8} -$10.8_{21.7}$\% & \cellcolor{red!8} -$9.8_{2.0}$\% & \cellcolor{red!5} -$2.3_{0.0}$\% & \cellcolor{red!9} -$14.1_{8.6}$\% & \cellcolor{red!11} -$19.7_{12.0}$\% & \cellcolor{green!13} +$0.5_{0.0}$\%\\
0.0001 & \cellcolor{red!10} -$15.4_{2.6}$\% & \cellcolor{red!10} -$15.5_{2.6}$\% & \cellcolor{red!5} -$0.4_{0.0}$\% & \cellcolor{red!12} -$22.6_{1.7}$\% & \cellcolor{red!12} -$22.4_{1.7}$\% & \cellcolor{green!9} +$0.3_{0.0}$\% & \cellcolor{red!12} -$22.4_{9.6}$\% & \cellcolor{red!13} -$24.1_{4.1}$\% & \cellcolor{red!5} -$2.6_{2.6}$\% & \cellcolor{red!13} -$22.9_{8.6}$\% & \cellcolor{red!15} -$30.4_{18.8}$\% & \cellcolor{red!6} -$3.1_{0.0}$\%\\
\bottomrule
\end{tabular}
}
\caption{Relative change in accuracy (\%), averaged over 3 seeds (mean $\pm$ std), for pruning with Llama 8B and RACE.}
\label{tab:llama8b_pruning_results_RACE}
\end{table*}

\begin{table*}[h]
\centering

\resizebox{\textwidth}{!}{
\begin{tabular}{l ccc ccc ccc ccc}
\toprule
\multicolumn{1}{l}{\multirow{2}{*}{Top-$k$}} & \multicolumn{3}{c}{English} & \multicolumn{3}{c}{German} & \multicolumn{3}{c}{Hindi} & \multicolumn{3}{c}{French} \\
\cmidrule(rl){2-4} \cmidrule(rl){5-7} \cmidrule(rl){8-10} \cmidrule(rl){11-13}
 & GSM8K & GSM8K flex & MMLU & GSM8K & GSM8K flex & MMLU & GSM8K & GSM8K flex & MMLU & GSM8K & GSM8K flex & MMLU \\
\hline
0.0 \small(Pre-train) & 0.765 & 0.775 & 0.679 & 0.585 & 0.590 & 0.538 & 0.415 & 0.490 & 0.450 & 0.580 & 0.585 & 0.512\\
\hline
0.000001 & \cellcolor{red!9} -$12.0_{6.5}$\% & \cellcolor{red!9} -$13.2_{5.2}$\% & \cellcolor{red!5} -$0.9_{0.0}$\% & \cellcolor{green!40} +$1.2_{0.0}$\% & \cellcolor{green!29} +$0.8_{1.7}$\% & +$0.0_{0.0}$\% & \cellcolor{red!14} -$27.7_{4.8}$\% & \cellcolor{red!7} -$6.7_{4.1}$\% & \cellcolor{green!30} +$0.9_{0.0}$\% & \cellcolor{red!9} -$11.6_{5.2}$\% & \cellcolor{red!9} -$14.0_{8.5}$\% & \cellcolor{red!5} -$0.2_{0.0}$\%\\
0.00001 & \cellcolor{red!6} -$4.8_{2.6}$\% & \cellcolor{red!6} -$5.4_{2.6}$\% & \cellcolor{red!5} -$0.4_{0.0}$\% & +$0.0_{1.7}$\% & \cellcolor{red!5} -$0.5_{0.0}$\% & \cellcolor{red!5} -$0.4_{0.0}$\% & \cellcolor{red!10} -$15.7_{19.3}$\% & \cellcolor{red!7} -$7.1_{0.0}$\% & \cellcolor{green!18} +$0.4_{0.0}$\% & \cellcolor{red!8} -$11.2_{0.0}$\% & \cellcolor{red!9} -$11.6_{1.7}$\% & \cellcolor{red!5} -$0.4_{0.0}$\%\\
0.0001 & \cellcolor{red!9} -$14.0_{1.3}$\% & \cellcolor{red!10} -$14.8_{1.3}$\% & \cellcolor{red!5} -$2.5_{1.5}$\% & \cellcolor{red!9} -$14.2_{3.4}$\% & \cellcolor{red!10} -$14.4_{5.1}$\% & \cellcolor{red!5} -$0.6_{0.0}$\% & \cellcolor{red!11} -$18.6_{2.4}$\% & \cellcolor{red!12} -$20.4_{4.1}$\% & \cellcolor{red!5} -$0.7_{0.0}$\% & \cellcolor{red!14} -$26.4_{3.4}$\% & \cellcolor{red!16} -$32.0_{10.3}$\% & \cellcolor{red!5} -$1.4_{0.0}$\%\\
\bottomrule
\end{tabular}
}
\caption{Relative change in accuracy (\%), averaged over 3 seeds (mean $\pm$ std), for pruning with Llama 8B and MMLU.}
\label{tab:llama8b_pruning_results_MMLU}
\end{table*}

\subsection{Collective vs.\ Individual Effects}
We examine whether the identified math-specific parameters operate primarily through a collective mechanism or exert independent (individual) effects. For all experiments in this section, we set \texttt{max\_new\_tokens} to 300 across all languages and use RACE as the non-math dataset.

\begin{figure}[t]
    \centering
    \includegraphics[width=\linewidth]{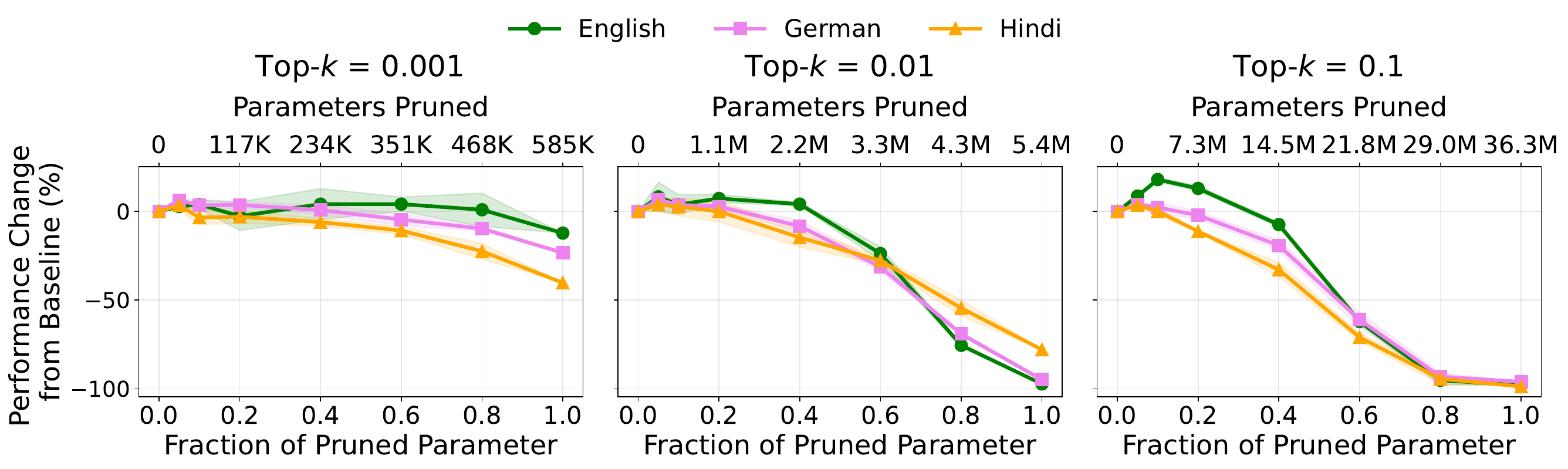}
    \caption{Effect of pruning isolated math-specific parameters on performance across languages for Qwen3 4B at top-$k \in \{0.001, 0.01, 0.1\}$.}
    \label{fig:qwen3_group_vs_individual}
\end{figure}

\begin{figure}[t]
    \centering
    \includegraphics[width=\linewidth]{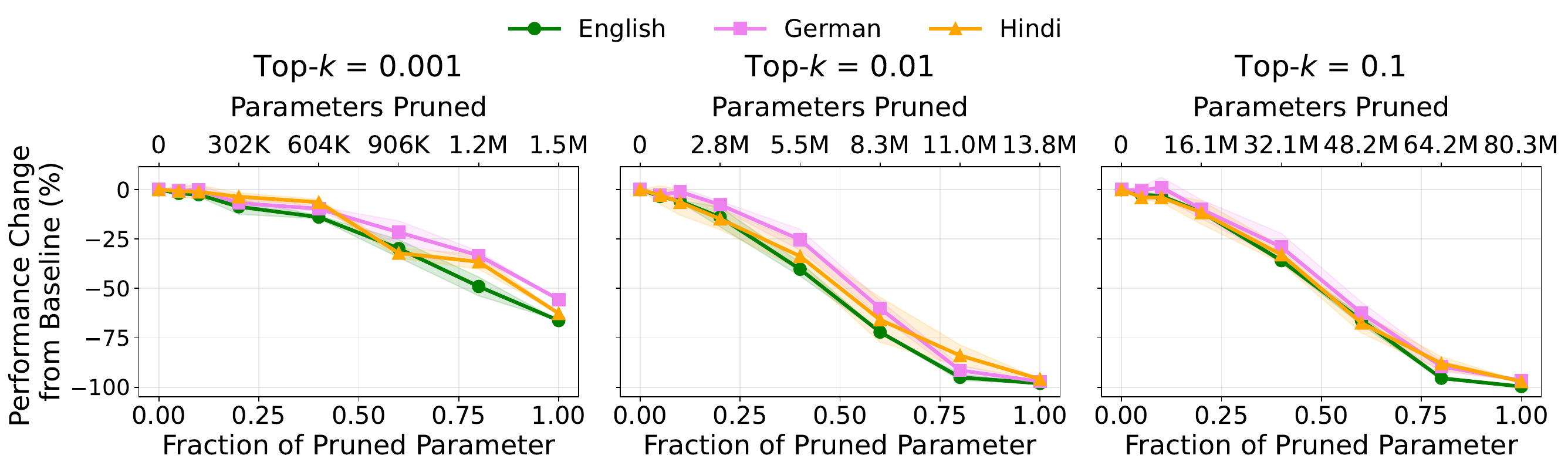}
\caption{Effect of pruning isolated math-specific parameters on performance across languages for Llama 8B at top-$k \in \{0.001, 0.01, 0.1\}$.}
    \label{fig:llama8b_group_vs_individual}
\end{figure}

As discussed in Section~\ref{sec:group_vs_individual} and shown in Figure~\ref{fig:llama1b_group_vs_individual}, the Llama 1B model exhibits collective behavior. This trend persists in larger models: Figures~\ref{fig:qwen3_group_vs_individual} and~\ref{fig:llama8b_group_vs_individual} show that both Qwen3 4B and Llama 8B demonstrate similar collective dynamics, suggesting that this phenomenon is consistent across model scales.

\section{Qualitative Examples for Scaling \& Pruning}
\setcounter{figure}{0}
\setcounter{table}{0}
\renewcommand{\thefigure}{\thesection.\arabic{figure}}
\renewcommand{\thetable}{\thesection.\arabic{table}}

In this section, we present several qualitative examples where scaling improves the model’s initial answers for mathematical reasoning across languages (Figures~\ref{fig:scaling_example_en}-\ref{fig:scaling_example_fr}).
Furthermore, we show examples where pruning steers the model toward incorrect answers for mathematical reasoning (Figures~\ref{fig:pruning_example_en}-\ref{fig:pruning_example_fr}), alongside a notable exception in which pruning instead improves performance (Figure~\ref{fig:qwen_pruning_improvement_example_en}), as discussed in Appendix~\ref{sec:appendix:scaling_pruning_results}.

\begin{figure*}[t]
\centering
\setlength{\tabcolsep}{3pt} 
\begin{tabularx}{\linewidth}{X}
\begin{inputbox}
\textbf{Q:} Toula went to the bakery and bought various types of pastries. She bought 3 dozen donuts which cost \$68 per dozen, 2 dozen mini cupcakes which cost \$80 per dozen, and 6 dozen mini cheesecakes for \$55 per dozen. How much was the total cost?
\textbf{A:}
\end{inputbox}
\end{tabularx}


\setlength{\tabcolsep}{3pt} 
\begin{tabularx}{\linewidth}{X X}
\begin{customoutputbox}[equal height group=en]{Output before scaling (wrong)}
3 dozen donuts cost $3 \times 68 = 204$ dollars.\\
2 dozen mini cupcakes cost $2 \times 80 = 160$ dollars.\\
6 dozen mini cheesecakes cost $6 \times 55 = 330$ dollars.\\
\textbf{204 + 160 + 330 is 734. The answer is 734.}
\end{customoutputbox}
&
\begin{customoutputbox}[equal height group=en]{Output after scaling (correct)}
3 dozen donuts cost $3 \times 68 = 204$ dollars.\\
2 dozen mini cupcakes cost $2 \times 80 = 160$ dollars.\\
6 dozen mini cheesecakes cost $6 \times 55 = 330$ dollars.\\
\textbf{204 + 160 + 330 is 694. The answer is 694.}
\end{customoutputbox}
\end{tabularx}

\caption{Illustration of arithmetic inconsistency before scaling and corrected output after scaling for English reasoning using Llama 1B.}
\label{fig:scaling_example_en}

\end{figure*}
\begin{figure*}[t]
\centering
\setlength{\tabcolsep}{3pt} 
\begin{tabularx}{\linewidth}{X}
\begin{inputbox}
\textbf{Q:} Gunter versucht, die Marmeladebohnen in einem Glas zu zählen. Er fragt seine Freunde, wie viele in dem Glas sind. Einer sagt 80. Ein anderer sagt 20 mehr als die Hälfte des ersten. Ein
Drittel sagt 25\% mehr als der erste. Wie hoch ist ihre durchschnittliche Vermutung? \textbf{A:}

\vspace{3pt}
\hrule
\vspace{3pt}

\textcolor{gray}{\textit{(English translation: Gunter is trying to count the jam beans in a jar. He asks his friends how many are in the jar. One says 80. Another says 20, more than half the first. A third say 25\% more than the first. What is their average guess?)}}

\end{inputbox}
\end{tabularx}


\setlength{\tabcolsep}{3pt} 
\begin{tabularx}{\linewidth}{X X}
\begin{customoutputbox}[equal height group=de]{Output before scaling (wrong)}
Der erste sagt 80. Ein anderer sagt 20 mehr als die
Hälfte des ersten, also 80 + 20 = 100.
Ein Drittel sagt 25\% mehr als der erste, also 80 +
(80 * 0,25) = 100. Die durchschnittliche
Vermutung ist 100.
Die Antwort ist 100.

\vspace{3pt}
\hrule
\vspace{3pt}

\textcolor{gray}{\textit{(English translation: The first person says 80. Another says 20 more than
half of the first, so 80 + 20 = 100.
One third says 25\% more than the first, \textbf{so 80 + (80 * 0.25) = 100. The average guess is 100.
The answer is 100.})}}

\end{customoutputbox}
&
\begin{customoutputbox}[equal height group=de]{Output after scaling (correct)}
Der erste sagt 80. Ein anderer sagt 20 mehr als die
Hälfte des ersten, also 80 + 20 = 100.
Ein Drittel sagt 25\% mehr als der erste, also 80 +
(80 * 0,25) = 100 + 20 = 120. Die
durchschnittliche Vermutung ist 80.
Die Antwort ist 80.

\vspace{3pt}
\hrule
\vspace{3pt}

\textcolor{gray}{\textit{(English translation: The first person says 80. Another says 20 more than
half of the first, so 80 + 20 = 100.
One third says 25\% more than the first, \textbf{so 80 + (80 * 0.25) = 100 + 20 = 120. The average guess is 80.
The answer is 80.})}}

\end{customoutputbox}
\end{tabularx}

\caption{Illustration of arithmetic inconsistency before scaling and corrected output after scaling for German reasoning using Llama 1B.}
\label{fig:scaling_example_de}

\end{figure*}
\begin{figure*}[t]
\centering
\setlength{\tabcolsep}{3pt} 
\begin{tabularx}{\linewidth}{X}
\begin{inputbox}
\textbf{Q:} \hi{मैरी ने एक चिकन भोजन का ऑर्डर दिया जिसकी कीमत 12 डॉलर है, 5 पैकेट दूध की कीमत 3 डॉलर है, 4 सेबों की कीमत 1.50 डॉलर है, और पिज्जा के कुछ डिब्बे हैं। मैरी ने कुल 50 डॉलर का भुगतान किया। अगर प्रत्येक बॉक्स की कीमत 8.50 डॉलर है तो मैरी ने पिज्जा के कितने डिब्बे मंगवाए?} \textbf{A:}

\vspace{3pt}
\hrule
\vspace{3pt}

\textit{\textcolor{gray}{(English translation: Mary ordered a chicken meal that costs \$12, 5 packets of milk that cost \$3, 4 apples that cost \$1.50, and some boxes of pizza. Mary paid a total of \$50. How many boxes of pizza did Mary order if each box costs \$8.50?)}}
\end{inputbox}
\end{tabularx}


\setlength{\tabcolsep}{3pt} 
\begin{tabularx}{\linewidth}{X X}
\begin{customoutputbox}[equal height group=hi]{Output before scaling (wrong)}

\hi{मैरी ने एक चिकन की कीमत 12 डॉलर के लिए 1 बॉक्स खरीदा। उसके पास 50 - 12 = 38 डॉलर बचे हैं। 5 पैकेट दूध की कीमत 3 डॉलर है, जिसकी कुल कीमत 5 × 3 = 15 डॉलर है। उसके पास 38 - 15 = 23 डॉलर बचे हैं। 4 सेबों की कीमत 1.50 डॉलर है, जिसकी कुल कीमत 4 × 1.50 = 6 डॉलर है। उसके पास 23 - 6 = 17 डॉलर बचे हैं। पिज्जा के कुछ डिब्बे की कीमत 8.50 डॉलर है, जिसकी कुल कीमत 17 / 8.50 = 1.96 डॉलर है। उसके पास 1.96 डॉलर बचे हैं। 1.96 / 8.50 = 0.23। उसके पास 0.23 डिब्बे हैं। उत्तर 0.23 है।}

\vspace{3pt}
\hrule
\vspace{3pt}

\textcolor{gray}{\textit{(English translation: Mary bought a box of chicken for \$12. She has \$50 - 12 = \$38 left. 5 packets of milk cost \$3, for a total cost of \$5 × 3 = \$15. She has \$38 - 15 = \$23 left. 4 apples cost \$1.50, for a total cost of 4 × 1.50 = \$6. She has \$23 - 6 = \$17 left. Some boxes of pizza cost \$8.50, for a total cost of \textbf{\$17 / 8.50 = \$1.96. She has \$1.96 left. 1.96 / 8.50 = 0.23. She has 0.23 boxes. The answer is 0.23.})}}
\end{customoutputbox}
&
\begin{customoutputbox}[equal height group=hi]{Output after scaling (correct)}
\hi{मैरी ने एक चिकन की कीमत 12 डॉलर के लिए 1 बॉक्स खरीदा। उसके पास 50 - 12 = 38 डॉलर बचे हैं। 5 पैकेट दूध की कीमत 3 डॉलर है, जिसकी कुल कीमत 5 × 3 = 15 डॉलर है। उसके पास 38 - 15 = 23 डॉलर बचे हैं। 4 सेबों की कीमत 1.50 डॉलर है, जिसकी कुल कीमत 4 × 1.50 = 6 डॉलर है। उसके पास 23 - 6 = 17 डॉलर बचे हैं। पिज्जा के कुछ डिब्बे की कीमत 8.50 डॉलर है, जिसकी कुल कीमत 17 / 8.50 = 2 डिब्बे हैं। उत्तर 2 है।}

\vspace{3pt}
\hrule
\vspace{3pt}

\textit{\textcolor{gray}{(English translation: Mary bought a box of chicken for \$12. She has \$50 - 12 = \$38 left. 5 packets of milk cost \$3, for a total cost of 5 × 3 = \$15. She has \$38 - 15 = \$23 left. 4 apples cost \$1.50, for a total cost of 4 × 1.50 = \$6. She has \$23 - 6 = \$17 left. Some boxes of pizza cost \$8.50, for a total cost of \textbf{17 / 8.50 = 2 boxes. The answer is 2.})}}
\end{customoutputbox}
\end{tabularx}

\caption{Illustration of arithmetic inconsistency before scaling and corrected output after scaling for Hindi reasoning using Llama 1B.}
\label{fig:scaling_example_hi}

\end{figure*}
\begin{figure*}[t]
\centering
\setlength{\tabcolsep}{3pt} 
\begin{tabularx}{\linewidth}{X}
\begin{inputbox}
\textbf{Q:} Peter prévoit d'aller au cinéma cette semaine. Il a toujours un billet à 7 \$ et du pop-corn à 7 \$. S'il a 42 dollars par semaine, combien de fois peut-il aller au cinéma ?\\
\textbf{A:}
\vspace{3pt}
\hrule
\vspace{3pt}

\textit{\textcolor{gray}{(English translation: Peter plans to go to the movies this week. He always has a \$7 ticket and \$7 popcorn. If he has \$42 a week, how many times can he go to the movies?)}}
\end{inputbox}
\end{tabularx}


\setlength{\tabcolsep}{3pt} 
\begin{tabularx}{\linewidth}{X X}
\begin{customoutputbox}[equal height group=fr_s]{Output before scaling (wrong)}
Peter a 42 dollars. Il a 7 \$ pour le billet et 7 \$ pour le pop-corn. Donc 42 - 7 - 7 = 28. Il peut aller au cinéma 28 fois. La réponse est 28.
\vspace{3pt}
\hrule
\vspace{3pt}

\textcolor{gray}{\textit{(English translation: Peter has \$42. He has \$7 for the ticket and \$7 for the popcorn. \textbf{So 42 - 7 - 7 = 28. He can go to the movies 28 times. The answer is 28.})}}
\end{customoutputbox}
&
\begin{customoutputbox}[equal height group=fr_s]{Output after scaling (correct)}
Peter a 42 dollars. 7 dollars pour le billet et 7 dollars pour le pop-corn font 7 + 7 = 14 dollars. Il peut donc aller au cinéma 42 / 14 = 3 fois. La réponse est 3.
\vspace{3pt}
\hrule
\vspace{3pt}

\textit{\textcolor{gray}{(English translation: Peter has \$42. \$7 for the ticket and \$7 for the popcorn make \textbf{\$7 + \$7 = \$14. He can therefore go to the movies 42 / 14 = 3 times. The answer is 3.})}}
\end{customoutputbox}
\end{tabularx}

\caption{Illustration of arithmetic inconsistency before scaling and corrected output after scaling for French reasoning using Llama 1B.}
\label{fig:scaling_example_fr}

\end{figure*}

\begin{figure*}[t]
\centering
\setlength{\tabcolsep}{3pt} 
\begin{tabularx}{\linewidth}{X}
\begin{inputbox}
\textbf{Q:} Janet’s ducks lay 16 eggs per day. She eats three for breakfast every morning and bakes muffins for her friends every day with four. She sells the remainder at the farmers' market daily for \$2 per fresh duck egg. How much in dollars does she make every day at the farmers' market? \\
\textbf{A:}
\end{inputbox}
\end{tabularx}


\setlength{\tabcolsep}{3pt} 
\begin{tabularx}{\linewidth}{X X}
\begin{customoutputbox}[equal height group=en]{Output before pruning (correct)}
Janet has 16 eggs per day. She eats 3 for breakfast. So she has 16 - 3 = 13 eggs left. She bakes muffins for 4 eggs. So she has 13 - 4 = 9 eggs left. She sells 9 eggs for \$2 each.\\
\textbf{9 * 2 = 18 dollars. The answer is 18.}
\end{customoutputbox}
&
\begin{customoutputbox}[equal height group=en]{Output after pruning (wrong)}
 Janet started with 16 eggs. She eats 3 for breakfast. She bakes 4 muffins. She sells 4 - 3 = 2 eggs at the market.\\
\textbf{16 - 3 is 8. The answer is 8.}
\end{customoutputbox}
\end{tabularx}

\caption{Illustration of correct output before pruning and wrong output after pruning for English reasoning (top-$k$=0.01) using Llama 1B.}
\label{fig:pruning_example_en}

\end{figure*}
\begin{figure*}[t]
\centering
\setlength{\tabcolsep}{3pt} 
\begin{tabularx}{\linewidth}{X}
\begin{inputbox}
\textbf{Q:} Janets Enten legen 16 Eier am Tag. Sie isst jeden Morgen drei zum Frühstück und backt mit vier täglich Muffins für ihre Freunde. Sie verkauft den Rest auf dem Bauernmarkt täglich für 2 Dollar pro frisches Enten-Ei. Wie viel Dollar verdient sie jeden Tag auf dem Bauernmarkt? \\
\textbf{A:}

\vspace{3pt}
\hrule
\vspace{3pt}

\textit{\textcolor{gray}{(English translation: Janet's ducks lay 16 eggs a day. She eats three for breakfast every morning and bakes muffins with four of them daily for her friends. She sells the rest at the farmers market every day for \$2 per fresh duck egg. How many dollars does she earn each day at the farmers market?)}}

\end{inputbox}
\end{tabularx}


\setlength{\tabcolsep}{3pt} 
\begin{tabularx}{\linewidth}{X X}
\begin{customoutputbox}[equal height group=de_prune]{Output before pruning (correct)}
Janets Enten legen 16 Eier am Tag. Sie isst 3 zum Frühstück und backt 4 Muffins. Das sind 3 + 4 = 7 Eier pro Tag. Sie verkauft 16 - 7 = 9 Eier pro Tag. Sie verdient 2 Dollar pro Ei.\\
9 * 2 = 18. Die Antwort ist 18.

\vspace{3pt}
\hrule
\vspace{3pt}

\textit{\textcolor{gray}{(English translation: Janet's ducks lay 16 eggs a day. She eats 3 for breakfast and bakes 4 muffins. That's 3 + 4 = 7 eggs per day. She sells 16 - 7 = 9 eggs per day. She earns \$2 per egg. \textbf{9 * 2 = 18. The answer is 18.})}}

\end{customoutputbox}
&
\begin{customoutputbox}[equal height group=de_prune]{Output after pruning (wrong)}
 Janets Enten legten 16 Eier am Tag. Sie isst jeden Morgen drei zum Frühstück und backt mit vier täglich Muffins für ihre Freunde. Sie verkauft den Rest auf dem Bauernmarkt täglich für 2 Dollar pro frisches Ei.\\
16 - 2 = 4. Die Antwort ist 4. Die Antwort ist 4. Die Antwort ist 4. Die Antwort ist 4. Die Antwort ist 4.

\vspace{3pt}
\hrule
\vspace{3pt}

\textit{\textcolor{gray}{(English translation: Janet's ducks laid 16 eggs a day. She eats three for breakfast every morning and bakes muffins for her friends with four of them daily. She sells the rest at the farmers market every day for \$2 a fresh egg. \textbf{16 - 2 = 4. The answer is 4. The answer is 4. The answer is 4. The answer is 4. The answer is 4.})}}

\end{customoutputbox}
\end{tabularx}

\caption{Illustration of correct output before pruning and wrong output after pruning for German reasoning (top-$k$=0.01)  using Llama 1B.}
\label{fig:pruning_example_de}

\end{figure*}

\begin{figure*}[t]
\centering
\setlength{\tabcolsep}{3pt} 
\begin{tabularx}{\linewidth}{X}
\begin{inputbox}
\textbf{Q:} 

\hi{हेनरी ने अपनी 60 मील की साइकिल यात्रा के दौरान दो बार रुका। वह पहली बार 20 मील के बाद रुका। यात्रा के अंत से 15 मील पहले उनका दूसरा पड़ाव था। उसने अपने पहले और दूसरे स्टॉप के बीच कितने मील की यात्रा की?}

\textbf{A:}

\vspace{3pt}
\hrule
\vspace{3pt}

\textit{\textcolor{gray}{(English translation: Henry stopped twice during his 60-mile bicycle trip. He first stopped after 20 miles. His second stop was 15 miles before the end of the trip. How many miles did he travel between his first and second stops?)}}

\end{inputbox}
\end{tabularx}


\setlength{\tabcolsep}{3pt} 
\begin{tabularx}{\linewidth}{X X}
\begin{customoutputbox}[equal height group=hi_prune]{Output before pruning (correct)}

\hi{हेनरी ने अपनी यात्रा के दौरान 60 मील की यात्रा की और दो बार रुका।}
\hi{पहली बार 20 मील की यात्रा के बाद, उसका दूसरा पड़ाव 15 मील की यात्रा के बाद था।}
\hi{60 - 20 - 15 = 25 मील। उत्तर 25 है।}

\vspace{3pt}
\hrule
\vspace{3pt}

\textit{\textcolor{gray}{(English translation: Henry traveled 60 miles during his journey and stopped twice. The first time was after traveling 20 miles, and his second stop was after traveling 15 miles. \textbf{60 - 20 - 15 = 25 miles. The answer is 25.})}}
\end{customoutputbox}
&
\begin{customoutputbox}[equal height group=hi_prune]{Output after pruning (wrong)}

\hi{हेनरी ने अपनी 60 मील की साइकिल यात्रा के दौरान दो बार रुका। पहली बार 20 मील के बाद, उसका दूसरा पड़ाव 15 मील के बाद हुआ । 60 - 20 = 40। 60 - 20 = 40। 60 - 20 = 40। उत्तर 40 है}

\vspace{3pt}
\hrule
\vspace{3pt}

\textit{\textcolor{gray}{(English translation: Henry stopped twice during his 60-mile bicycle ride. The first time was after 20 miles, and his second stop was after 15 miles. \textbf{60 - 20 = 40. 60 - 20 = 40. 60 - 20 = 40. The answer is 40.})}}

\end{customoutputbox}
\end{tabularx}

\caption{Illustration of correct output before pruning and wrong output after pruning for Hindi reasoning (top-$k$=0.01) using Llama 1B.}
\label{fig:pruning_example_hi}

\end{figure*}
\begin{figure*}[t]
\centering
\setlength{\tabcolsep}{3pt} 
\begin{tabularx}{\linewidth}{X}
\begin{inputbox}
\textbf{Q:} James décide de courir 3 sprints 3 fois par semaine. Il court 60 mètres à chaque sprint. Combien de mètres au total par semaine ?\\
\textbf{A:}
\vspace{3pt}
\hrule
\vspace{3pt}

\textit{\textcolor{gray}{(English translation: James decides to run 3 sprints 3 times a week. He runs 60 meters in each sprint. How many meters in total per week?)}}
\end{inputbox}
\end{tabularx}


\setlength{\tabcolsep}{3pt} 
\begin{tabularx}{\linewidth}{X X}
\begin{customoutputbox}[equal height group=fr]{Output before pruning (correct)}
James court 3 sprints 3 fois par semaine. Cela fait 3 × 3 = 9 sprints. Chaque sprint fait 60 mètres. Donc 9 × 60 = 540 mètres. La réponse est 540.
\vspace{3pt}
\hrule
\vspace{3pt}

\textcolor{gray}{\textit{(English translation: James runs 3 sprints 3 times a week. That makes 3 × 3 = 9 sprints. Each sprint is 60 meters. \textbf{So 9 × 60 = 540 meters. The answer is 540.})}}
\end{customoutputbox}
&
\begin{customoutputbox}[equal height group=fr]{Output after pruning (wrong)}
James a 3 sprints 3 fois par semaine. Il court 60 mètres à chaque sprint. 3 + 3 = 5 sprints. 3 + 3 = 5. La réponse est 5.
\vspace{3pt}
\hrule
\vspace{3pt}

\textit{\textcolor{gray}{(English translation: James has 3 sprints 3 times a week. He runs 60 meters in each sprint. \textbf{3 + 3 = 5 sprints. 3 + 3 = 5. The answer is 5.})}}
\end{customoutputbox}
\end{tabularx}

\caption{Illustration of correct output before pruning and wrong output after pruning for French reasoning (top-$k$=0.01) using Llama 1B.}
\label{fig:pruning_example_fr}

\end{figure*}

\begin{figure*}[t]
\centering
\setlength{\tabcolsep}{3pt} 
\begin{tabularx}{\linewidth}{X}
\begin{inputbox}
Q: There are 15 trees in the grove. Grove workers 
will plant trees in the grove today. After they are 
done, there will be 21 trees. How many trees did the 
grove workers plant today?

A: There are 15 trees originally. Then there were 21 
trees after some more were planted. So there must 
have been 21 - 15 = 6. The answer is 6.

\vspace{2ex}
\texttt{[...]}
\vspace{2ex}



\textbf{Q:} A treasure hunter found a buried treasure chest filled with gems. There were 175 diamonds, 35 fewer rubies than diamonds, and twice the number of emeralds than the rubies.   How many of the gems were there in the chest?

\textbf{A:}

\end{inputbox}
\end{tabularx}


\setlength{\tabcolsep}{3pt} 
\begin{tabularx}{\linewidth}{X X}
\begin{customoutputbox}[equal height group=en_qwen]{Output before pruning (wrong format)}
 Let's break down the problem step by step:

1. **Number of diamonds**:  
   The problem states there are 175 diamonds.  
   So, diamonds = 175.

2. **Number of rubies**:  
   There are 35 fewer rubies than diamonds.  
   So, rubies = diamonds - 35 = 175 - 35 = 140.

3. **Number of emeralds**:  
   There are twice as many emeralds as rubies.  
   So, emeralds = 2 × rubies = 2 × 140 = 280.

4. **Total number of gems**:  
   Total gems = diamonds + rubies + emeralds  
   Total gems = 175 + 140 + 280 = 595.

**Answer: 595**. { \emoji ✅}

The total number of gems in the chest is **595**. { \emoji ✅}

Final Answer: **595**. {\emoji ✅}

\end{customoutputbox}
&
\begin{customoutputbox}[equal height group=en_qwen]{Output after pruning (correct format)}
The number of rubies is 175 - 35 = 140. The number of emeralds is 2 * 140 = 280. The total number of gems is 175 + 140 + 280 = 595. The answer is 595.
 
\end{customoutputbox}
\end{tabularx}

\caption{Illustration of wrong output formatting and poor in-context learning before pruning and correct output after pruning for English reasoning (top-$k$=1e-5) using Qwen3 4B.}
\label{fig:qwen_pruning_improvement_example_en}

\end{figure*}

\section{Similarities across Languages \& Tasks}
\label{sec:apeendix:languages_tasks_jaccard}
\setcounter{figure}{0}
\setcounter{table}{0}
\renewcommand{\thefigure}{\thesection.\arabic{figure}}
\renewcommand{\thetable}{\thesection.\arabic{table}}

In this section, we present plots showing the number of isolated parameters across languages (English (En), German (De), Hindi (Hi), and French (Fr)), as well as Jaccard similarity plots for each language pair, across all top-$k$ values and models.

\begin{figure*}
    \centering
    \includegraphics[width=\linewidth]{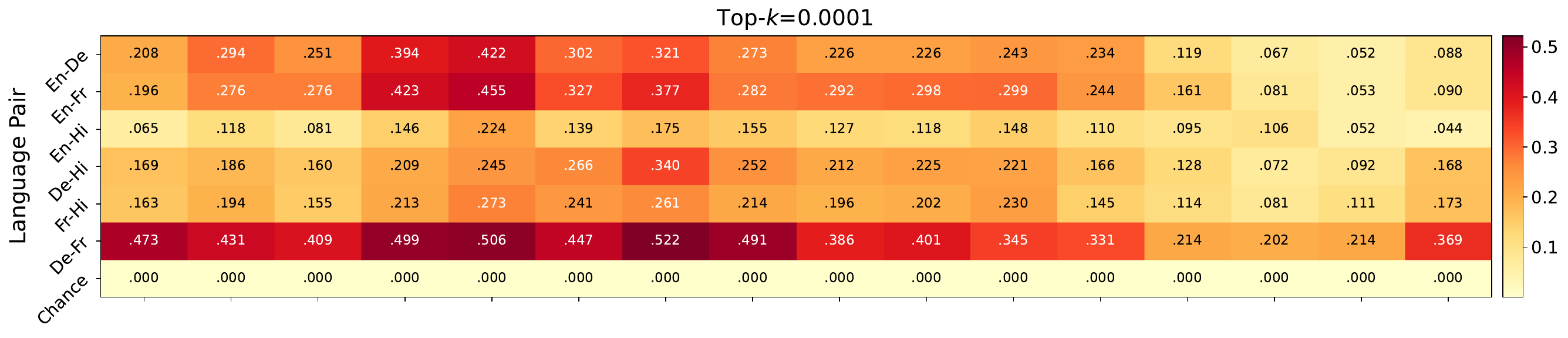}
    \includegraphics[width=\linewidth]{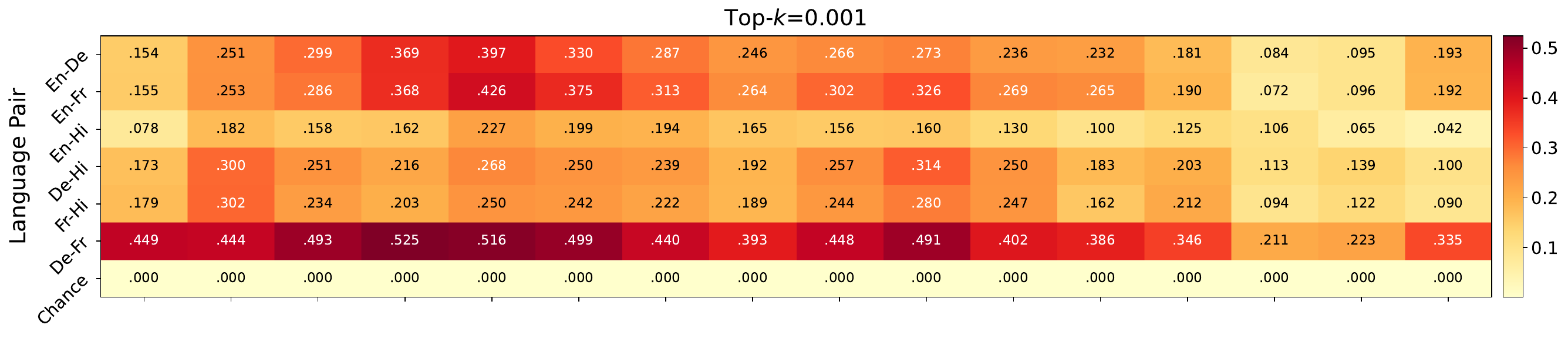}
    \includegraphics[width=\linewidth]{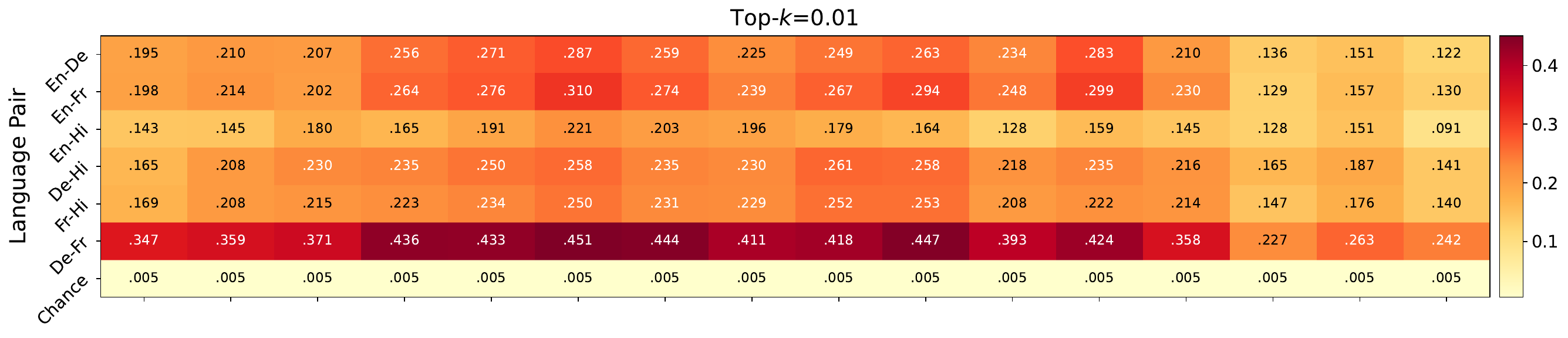}
    \includegraphics[width=\linewidth]{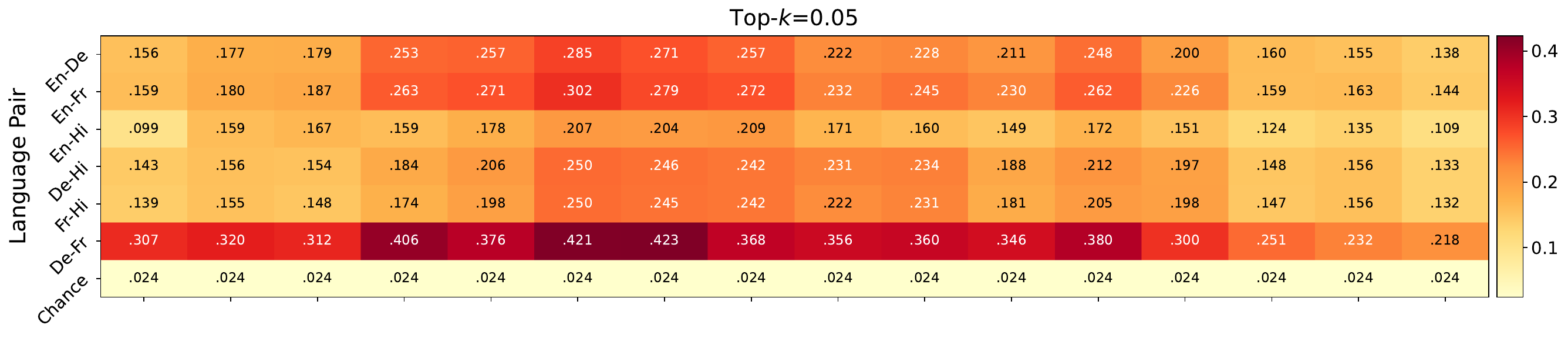}
    \includegraphics[width=\linewidth]{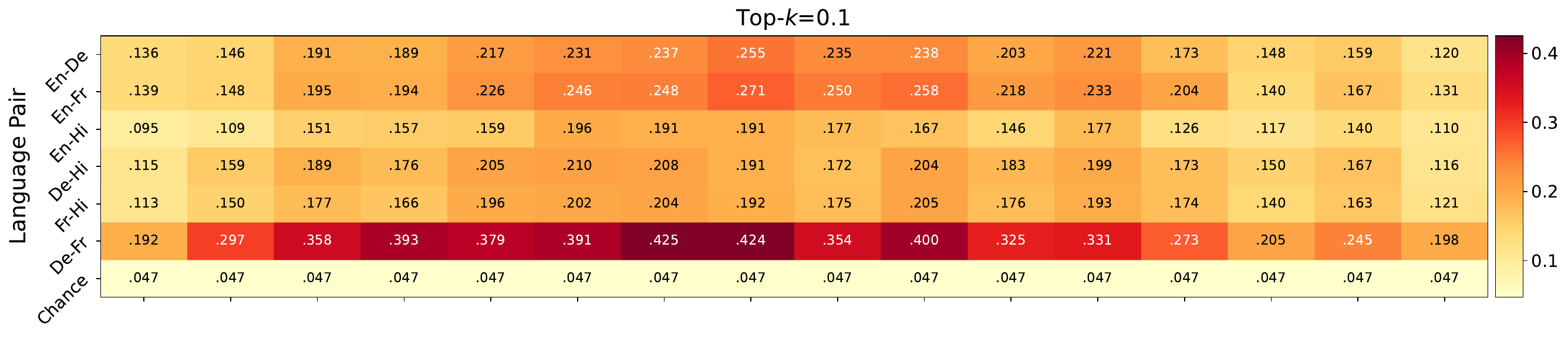}
    \includegraphics[width=\linewidth]{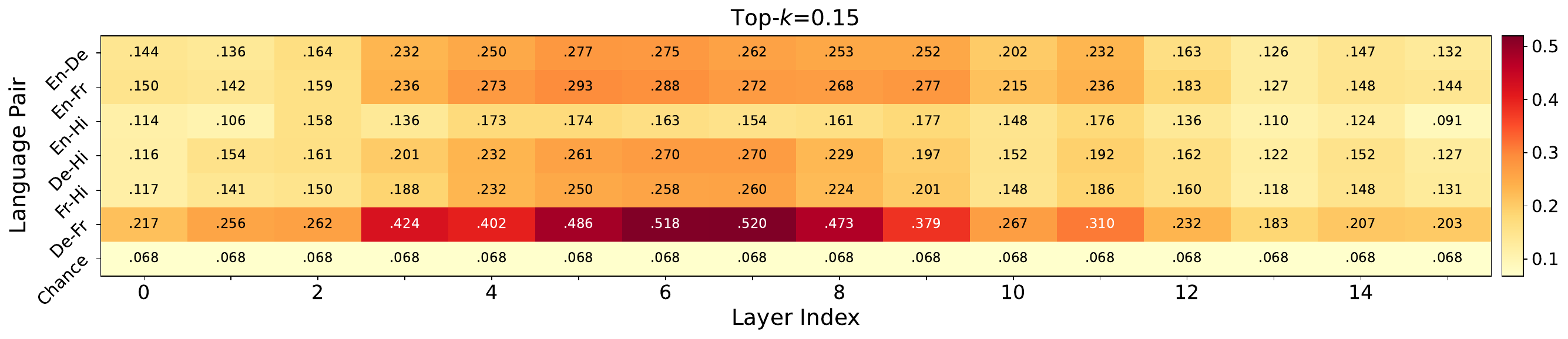}
\caption{Layer-wise Jaccard similarity of math-reasoning parameters across top-$k$ values in Llama 1B, computed for all language pairs (English (En), German (De), French (Fr), and Hindi (Hi)) and the random baseline (Chance) on RACE.}
     \label{fig:llama1b_all_topk_heatmap}
\end{figure*}

\begin{figure*}
    \centering
    \includegraphics[width=\linewidth]{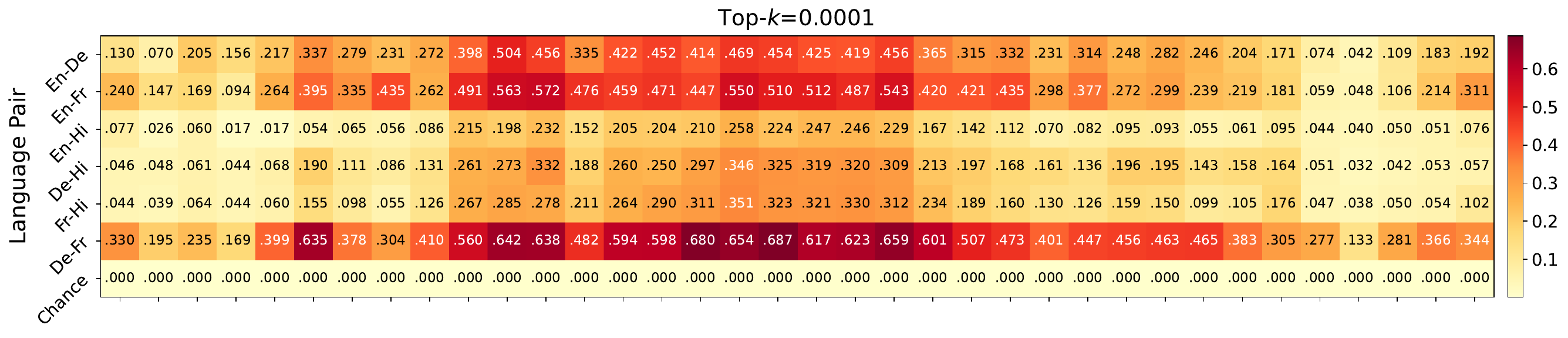}
    \includegraphics[width=\linewidth]{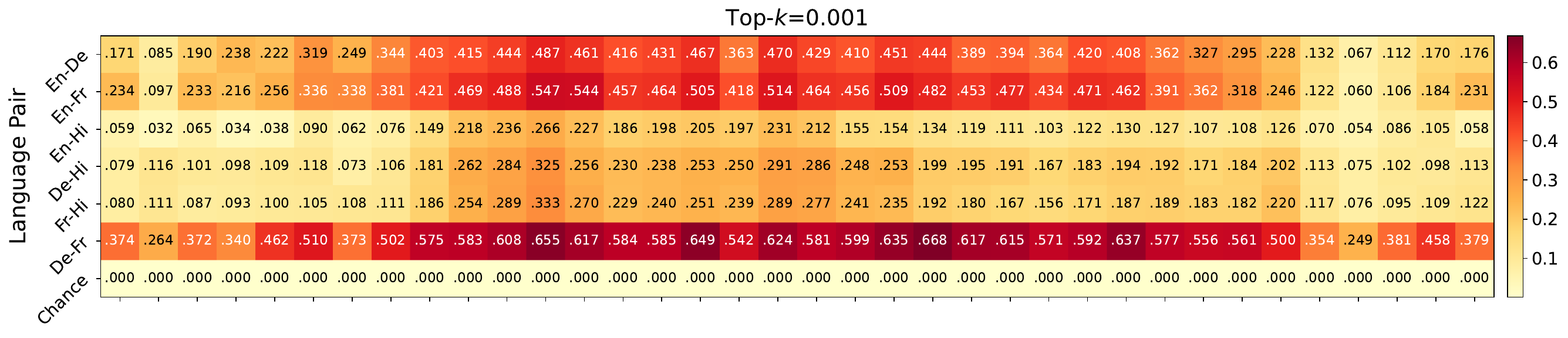}
    \includegraphics[width=\linewidth]{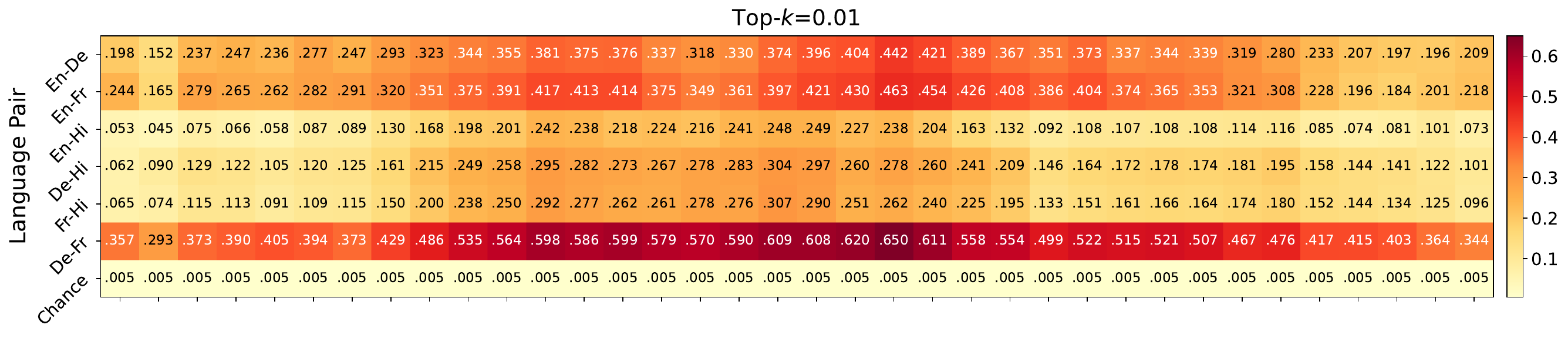}
    \includegraphics[width=\linewidth]{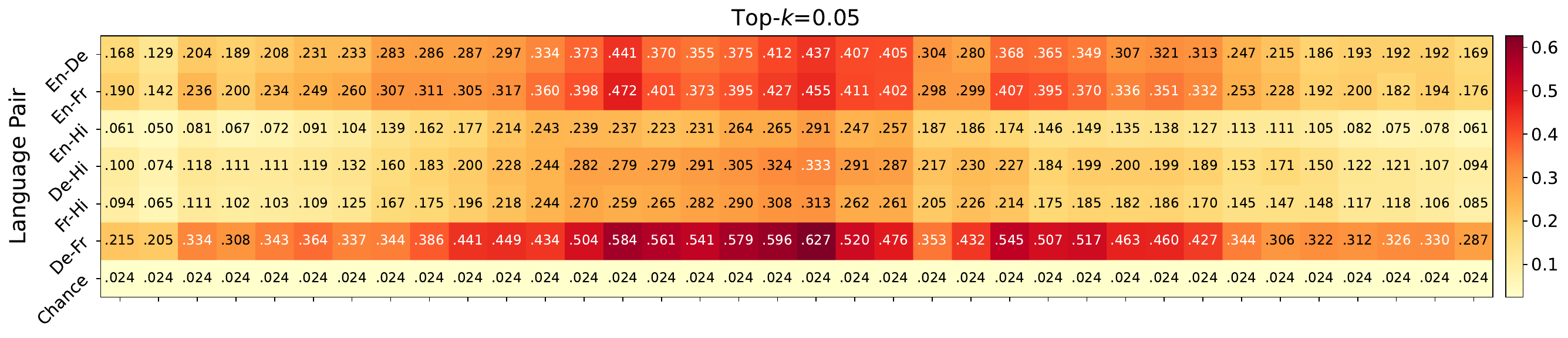}
    \includegraphics[width=\linewidth]{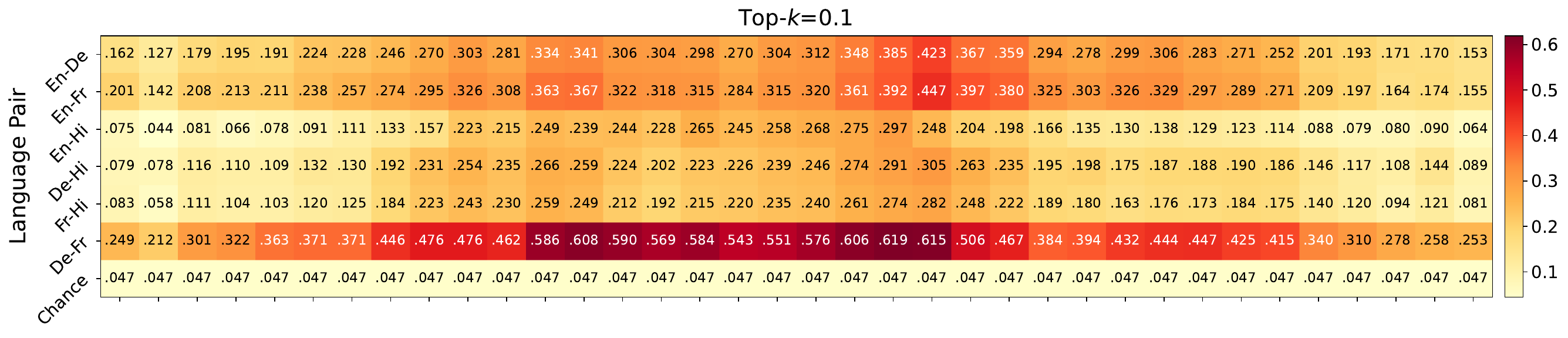}
    \includegraphics[width=\linewidth]{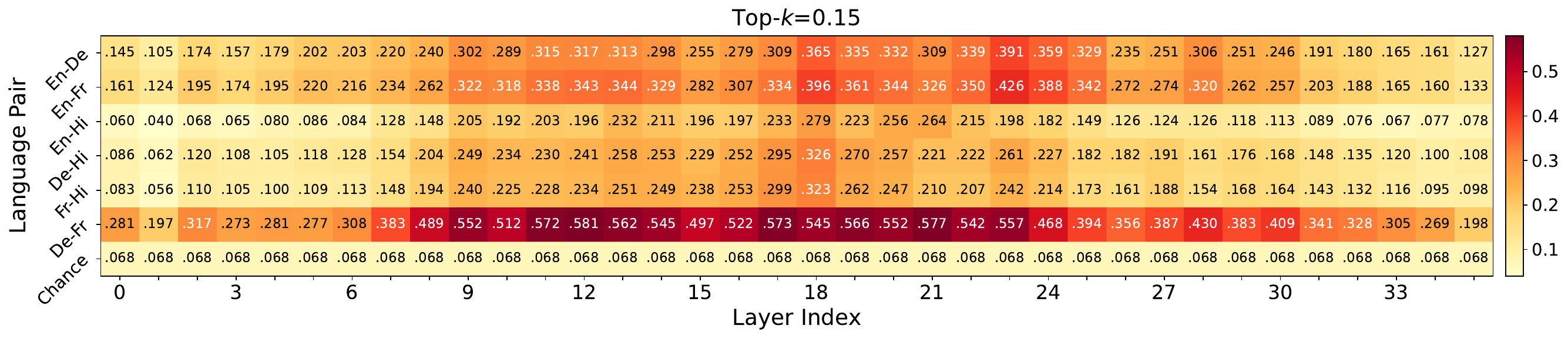}
    \caption{Layer-wise Jaccard similarity of math-reasoning parameters across top-$k$ values in Qwen3 4B, computed for all language pairs (English (En), German (De), French (Fr), and Hindi (Hi)) and the random baseline (Chance) on RACE.}
     \label{fig:Qwen3-4B_all_topk_heatmap}
\end{figure*}

\begin{figure*}
    \centering
    \includegraphics[width=\linewidth]{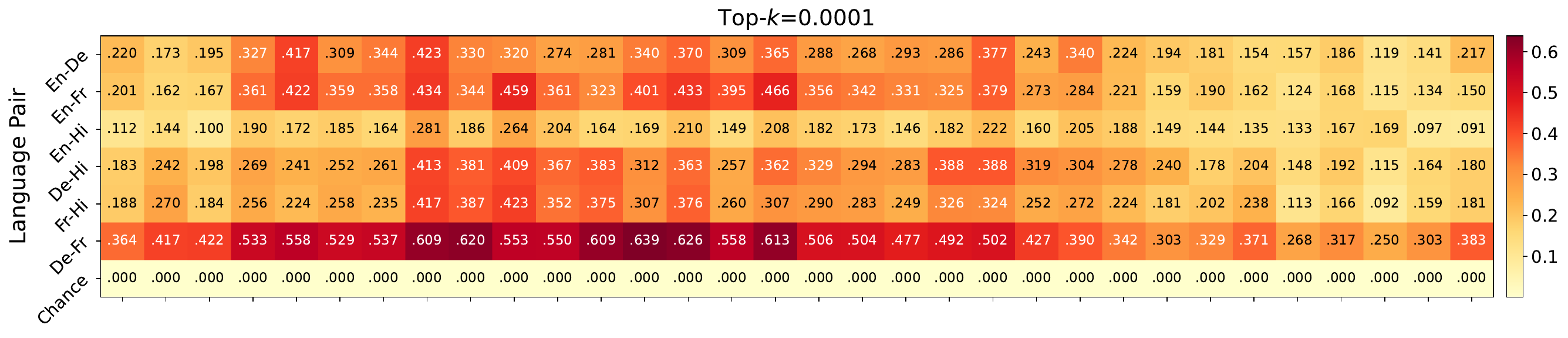}
    \includegraphics[width=\linewidth]{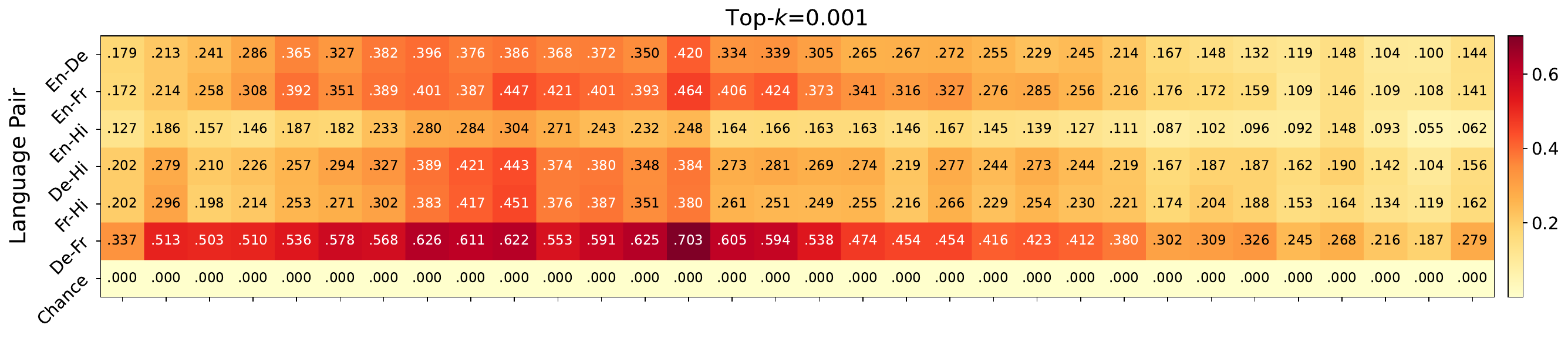}
    \includegraphics[width=\linewidth]{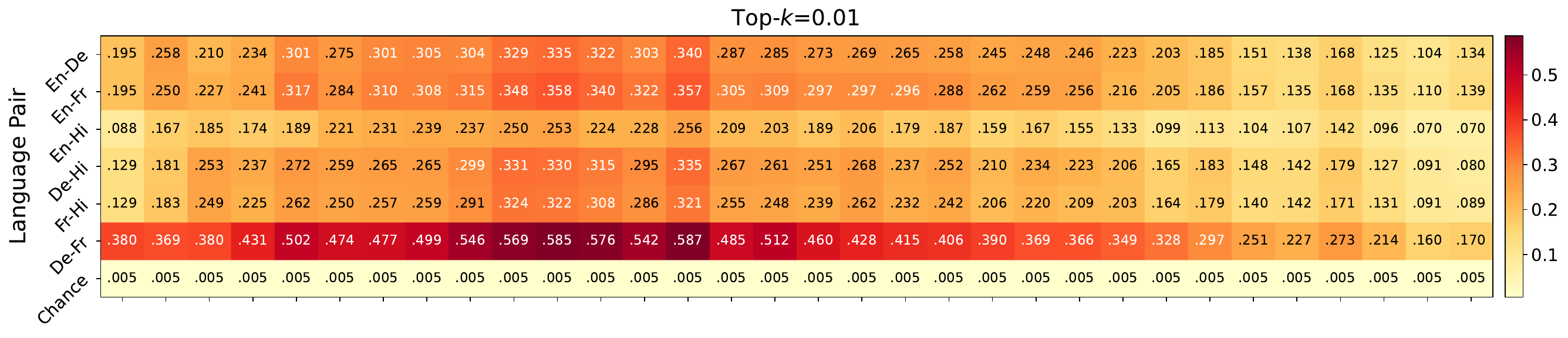}
    \includegraphics[width=\linewidth]{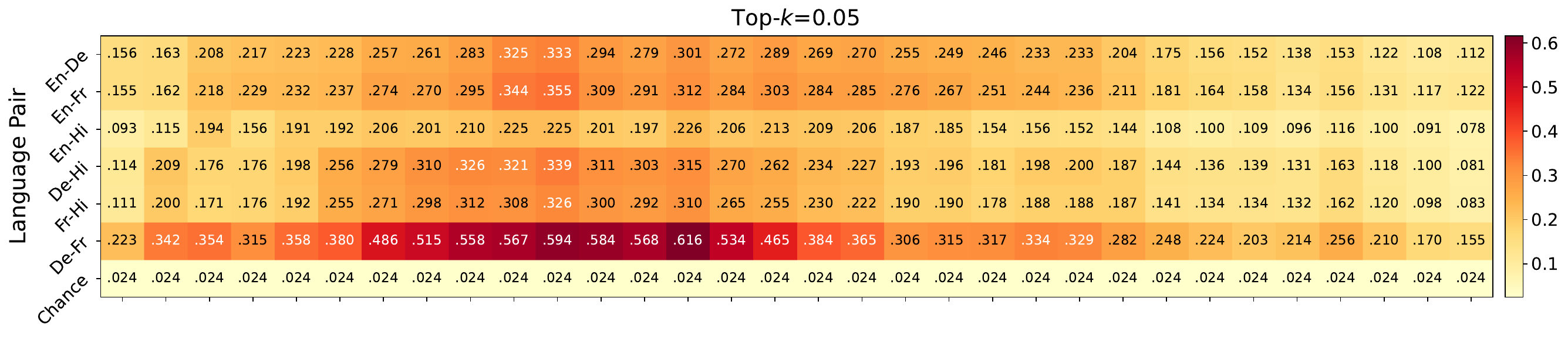}
    \includegraphics[width=\linewidth]{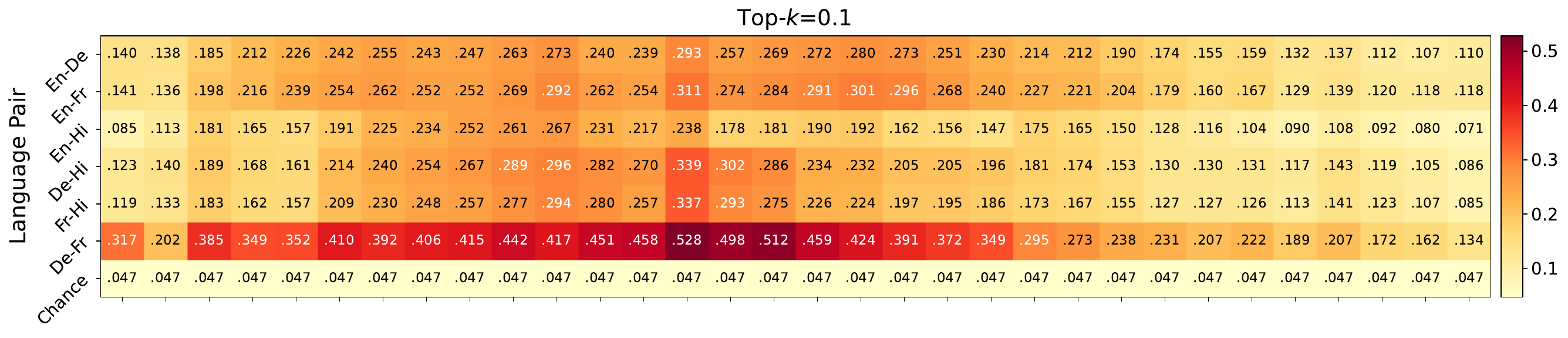}
    \includegraphics[width=\linewidth]{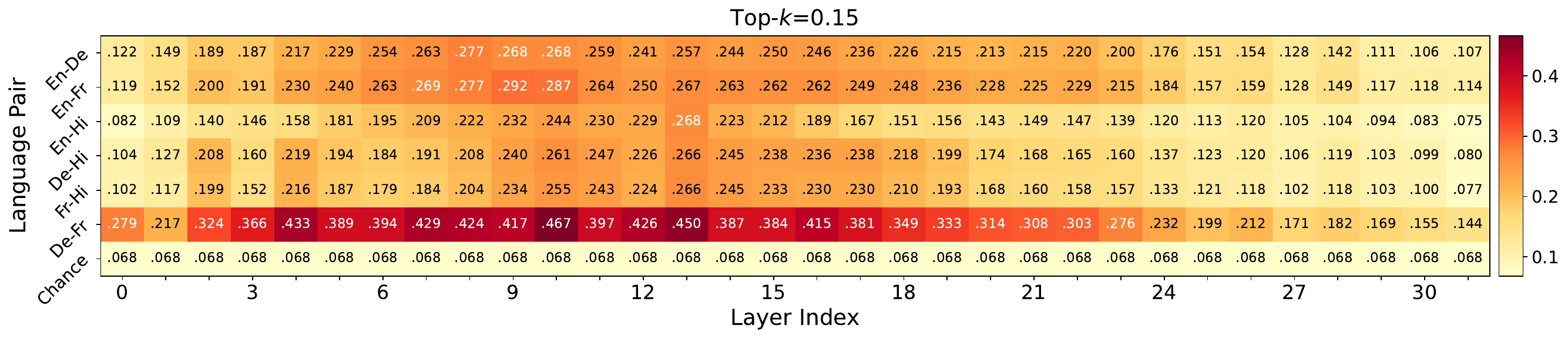}
    \caption{Layer-wise Jaccard similarity of math-reasoning parameters across top-$k$ values in Llama 8B, computed for all language pairs (English (En), German (De), French (Fr), and Hindi (Hi)) and the random baseline (Chance) on RACE.}
     \label{fig:Llama-8B_all_topk_heatmap}
\end{figure*}



\begin{figure*}[t]
    \centering
    \includegraphics[width=0.8\linewidth]{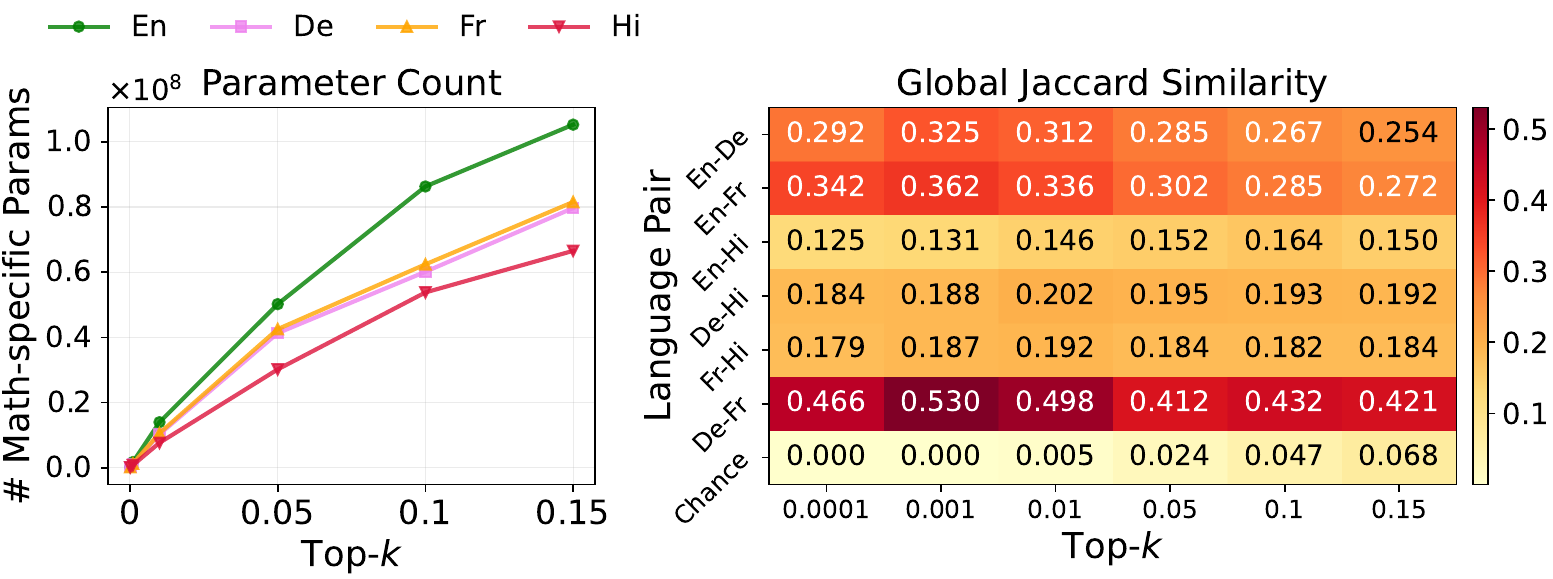}
\caption{Math-specific parameters in Qwen3 4B.
\textbf{Left}: Number of isolated parameters per language (English (En), German (De), French (Fr), and Hindi (Hi)).
\textbf{Right}: Jaccard similarity between language-specific parameter sets across all language pairs, compared to a random baseline using RACE as comparison dataset.}
    \label{fig:overall_jaccard_param_qwen4b}
\end{figure*}

\begin{figure*}[t]
    \centering
    \includegraphics[width=.8\linewidth]{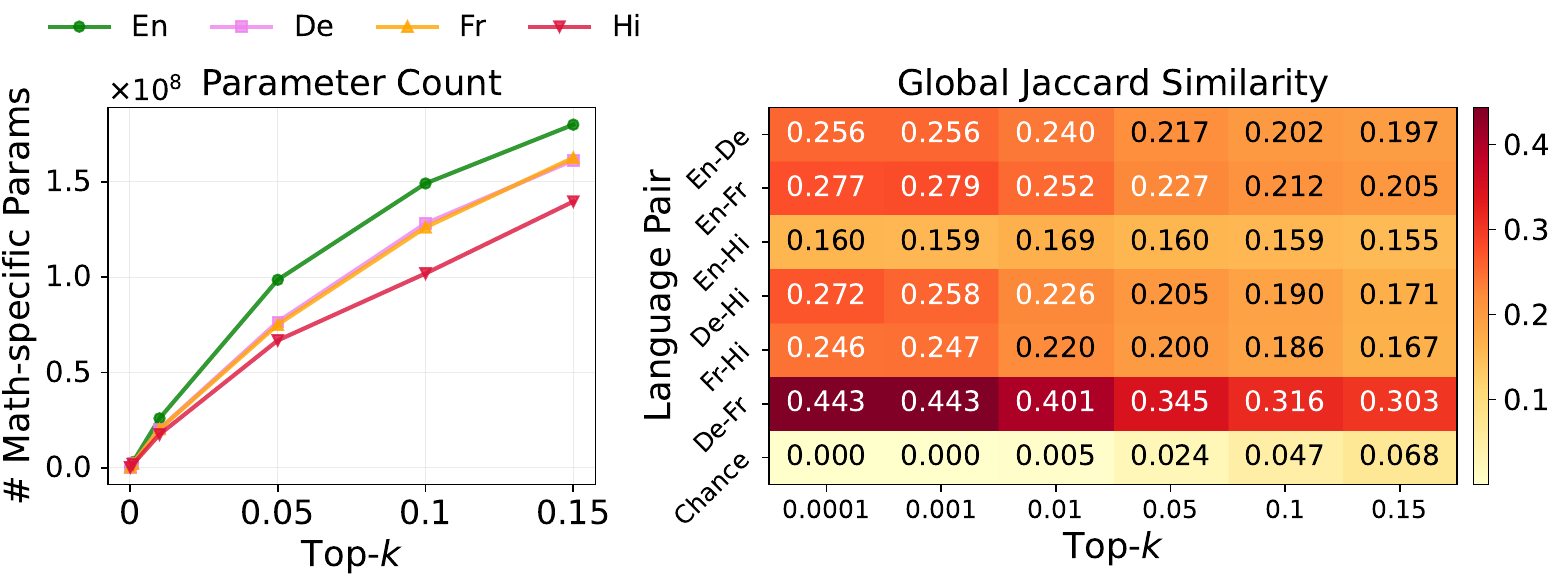}
    \caption{Math-specific parameters in Llama 8B.
\textbf{Left}: Number of isolated parameters per language (English (En), German (De), French (Fr), and Hindi (Hi)).
\textbf{Right}: Jaccard similarity between language-specific parameter sets across all language pairs, compared to a random baseline using RACE as comparison dataset.}
    \label{fig:overall_jaccard_param_llama8b}
\end{figure*}

\begin{figure*}[t]
    \centering
    \includegraphics[width=0.49\linewidth]{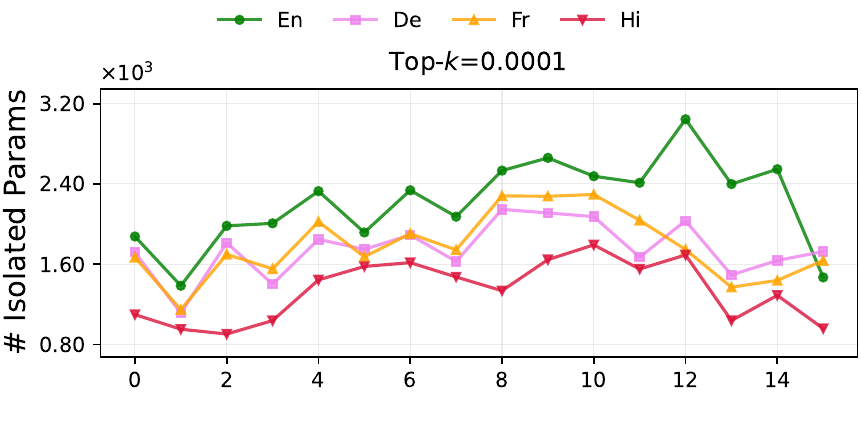}
    \includegraphics[width=0.49\linewidth]{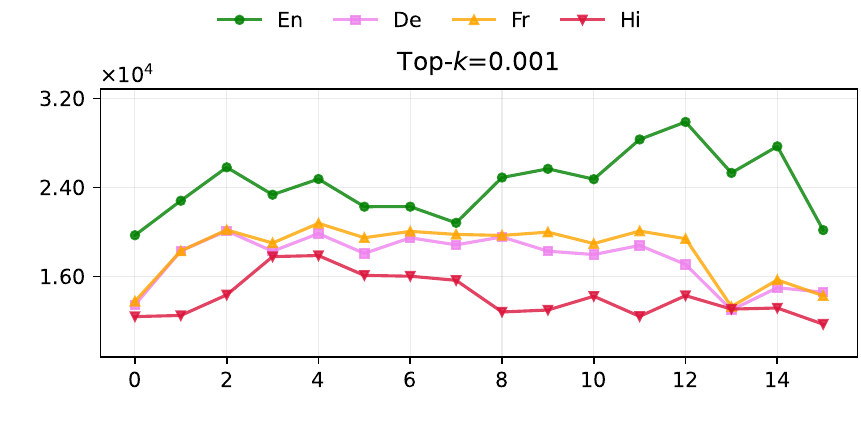}
    \includegraphics[width=0.49\linewidth]{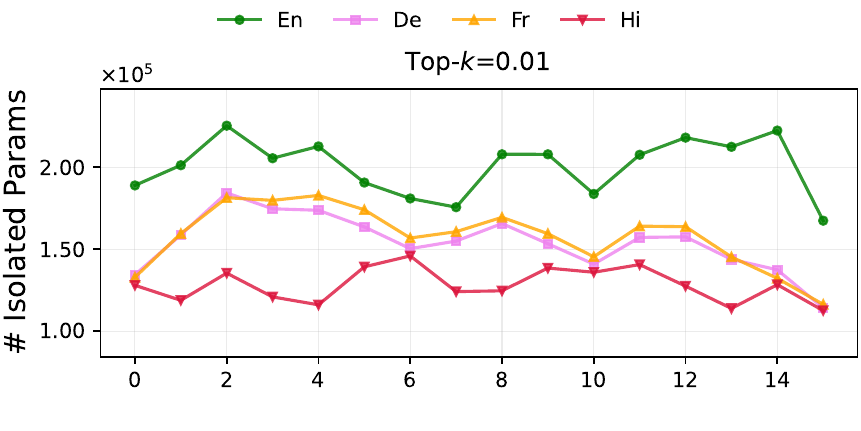}
    \includegraphics[width=0.49\linewidth]{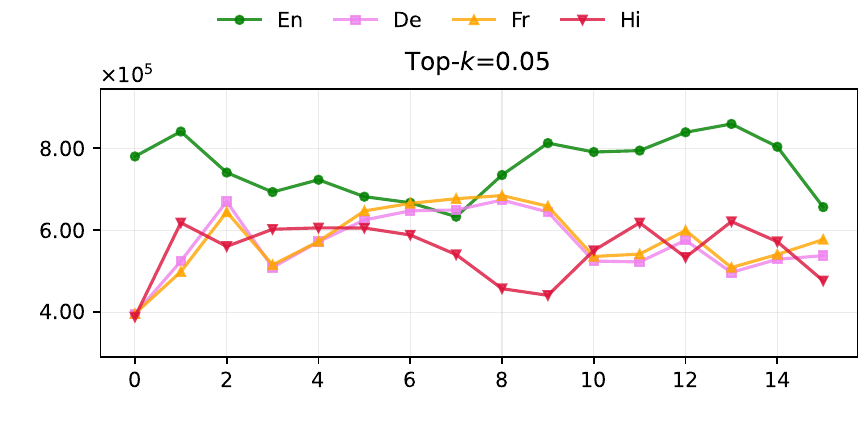}
    \includegraphics[width=0.49\linewidth]{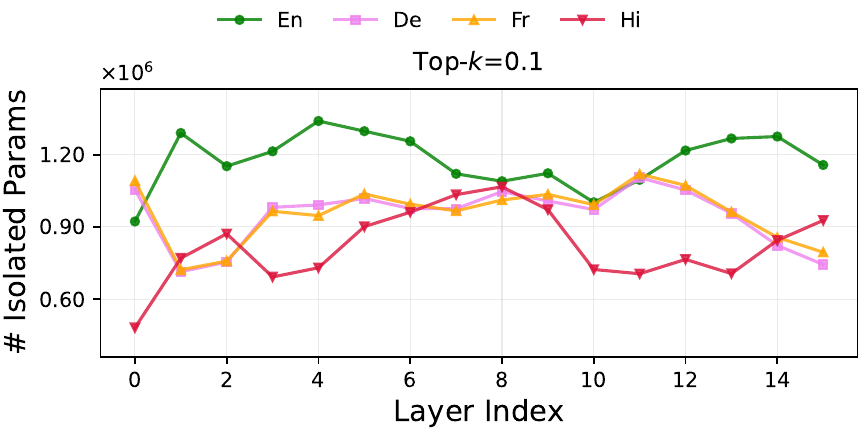}
    \includegraphics[width=0.49\linewidth]{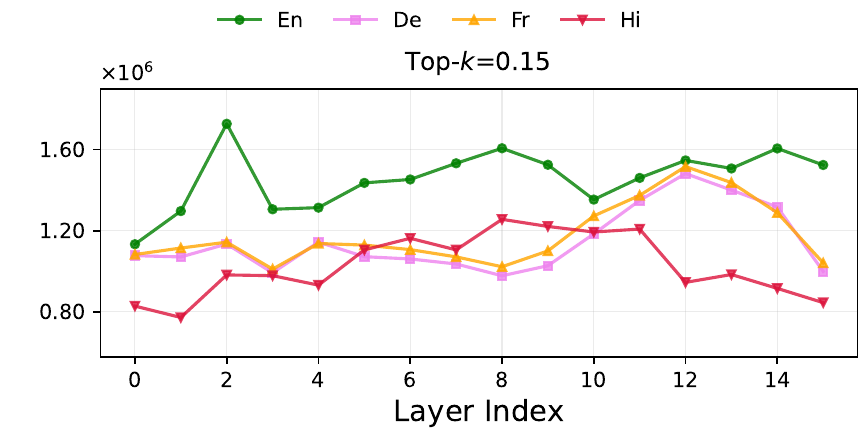}
    \caption{Layer-wise distribution of math-reasoning parameters across top-$k$ values in Llama 1B, using RACE as comparison dataset.}
    \label{fig:llama1b_layerwise_params_dist}
\end{figure*}
\begin{figure*}[t]
    \centering
    \includegraphics[width=0.49\linewidth]{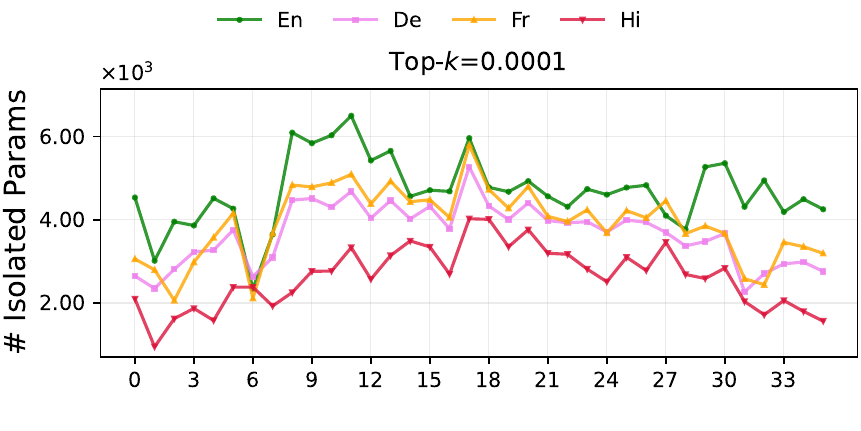}
    \includegraphics[width=0.49\linewidth]{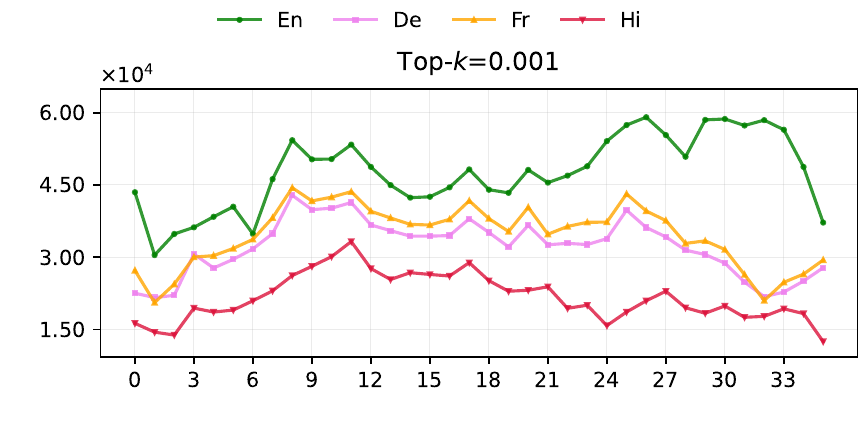}
    \includegraphics[width=0.49\linewidth]{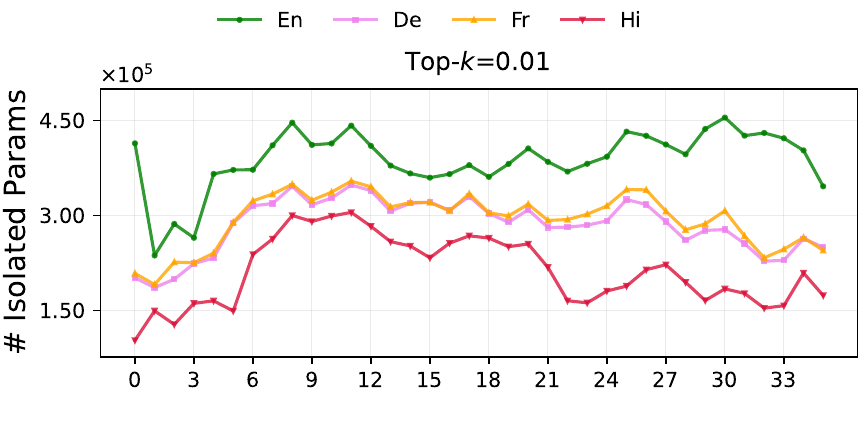}
    \includegraphics[width=0.49\linewidth]{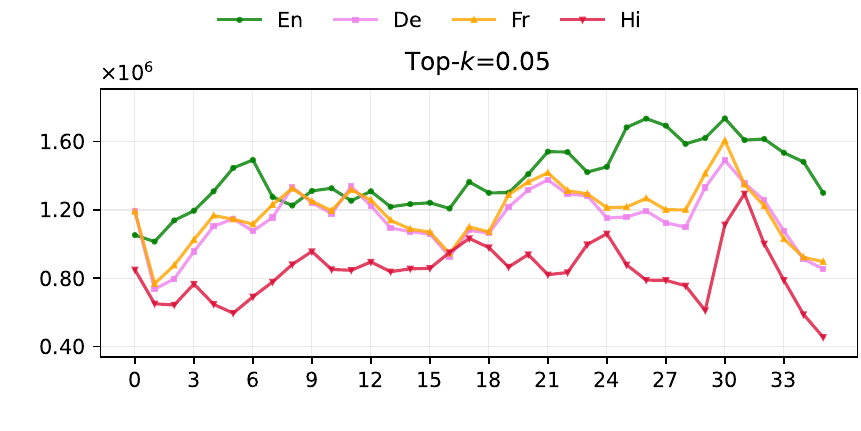}
    \includegraphics[width=0.49\linewidth]{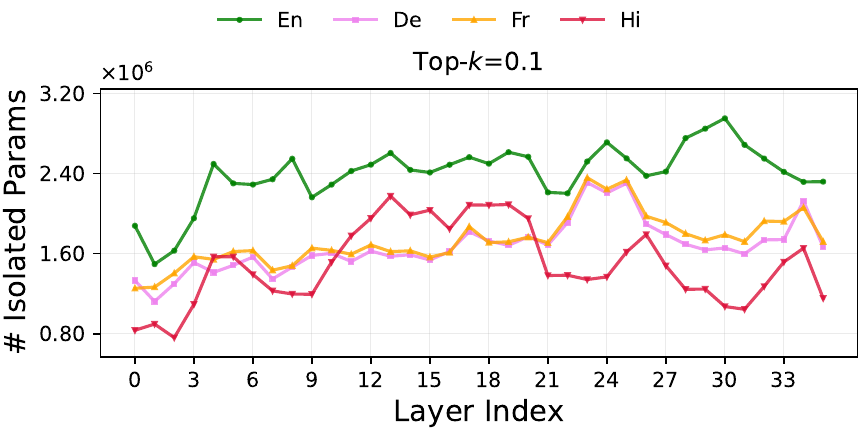}
    \includegraphics[width=0.49\linewidth]{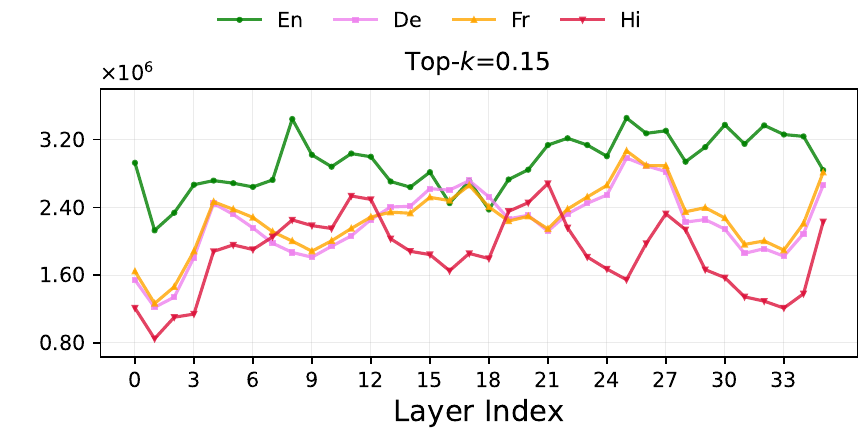}
    \caption{Layer-wise distribution of math-reasoning parameters across top-$k$ values in Qwen3 4B, using RACE as comparison dataset.}

    \label{fig:qwen4b_layerwise_params_dist}
\end{figure*}
\begin{figure*}[t]
    \centering
    \includegraphics[width=0.49\linewidth]{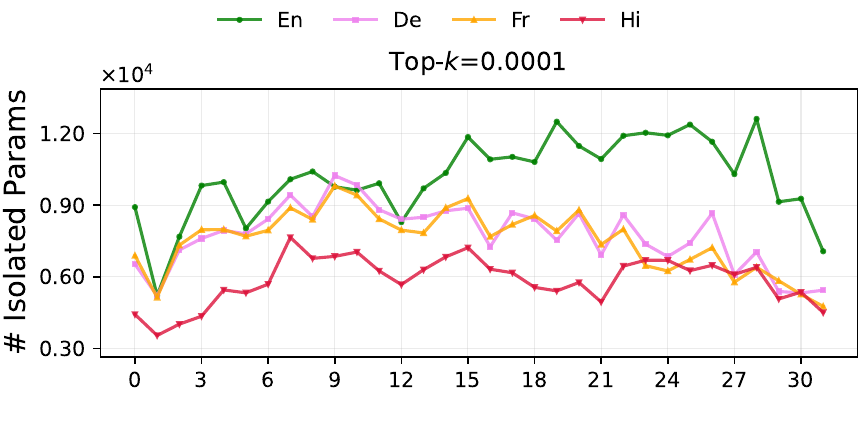}
    \includegraphics[width=0.49\linewidth]{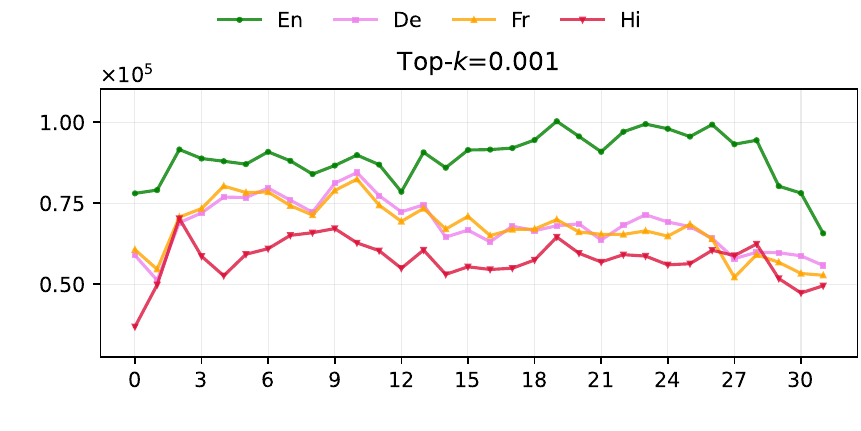}
    \includegraphics[width=0.49\linewidth]{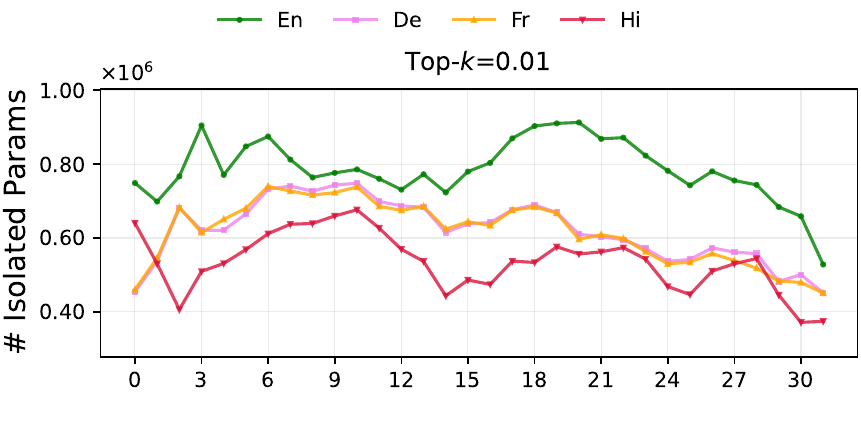}
    \includegraphics[width=0.49\linewidth]{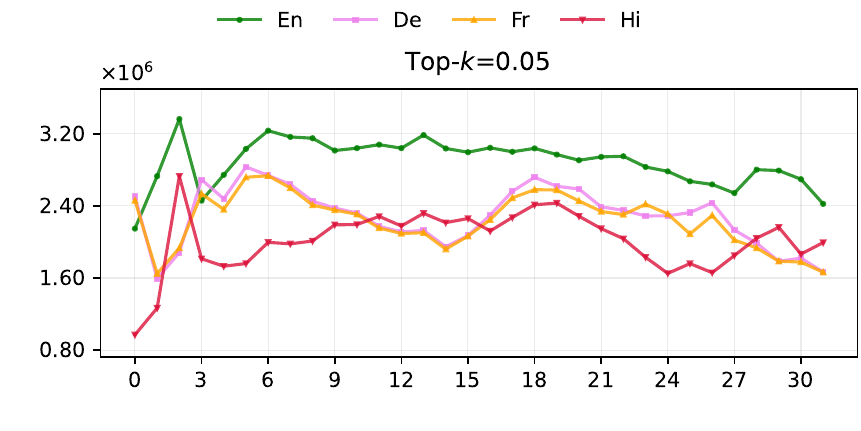}
    \includegraphics[width=0.49\linewidth]{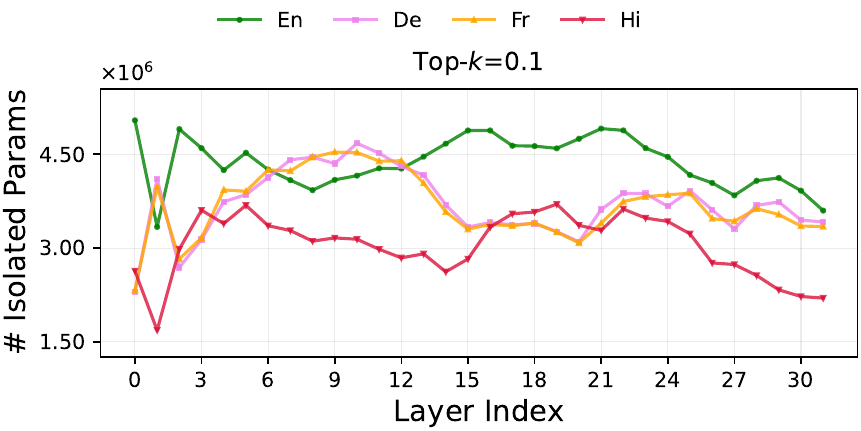}
    \includegraphics[width=0.49\linewidth]{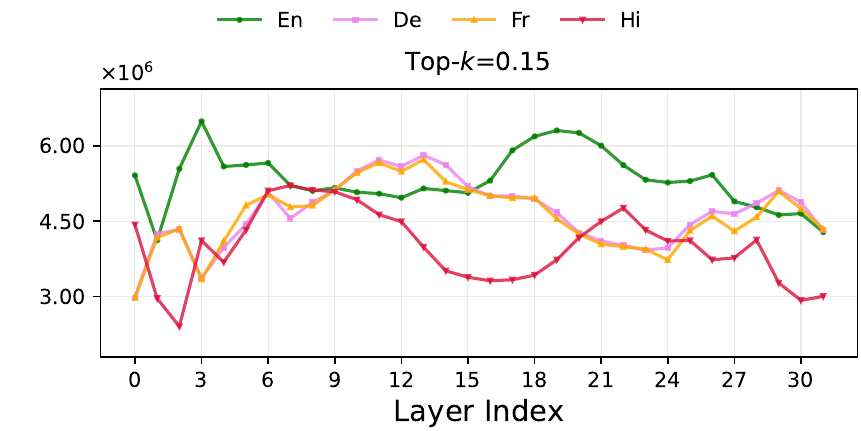}
    \caption{    Layer-wise distribution of math-reasoning parameters across top-$k$ values in Llama 8B, using RACE as comparison dataset.}
    \label{fig:llama8b_layerwise_params_dist}
\end{figure*}

\end{document}